\documentclass[AMS,STIX2COL]{WileyNJD-v2}

\articletype{Research Article}
\received{PREPRINT}
\revised{PREPRINT}
\accepted{PREPRINT}
\IfFileExists{upquote.sty}{\usepackage{upquote}}{}

\usepackage[utf8]{inputenc}
\usepackage{natbib}
\usepackage{graphicx}
\usepackage{amsmath}
\usepackage{amsthm} % for definition
\usepackage{xcolor}
\usepackage{hyperref}
\usepackage{movie15}
\usepackage[nottoc,numbib]{tocbibind} %references in table of contents

\newcommand\norm[1]{\lVert#1\rVert}

\begin{document}

\pagenumbering{arabic} 

% Title and authors
\title{Characterizing climate pathways using feature importance on echo state networks}
\author[1]{Katherine Goode$^{1}$*}
\author[1]{Daniel Ries$^{1}$}
\author[1]{Kellie McClernon$^{1}$}
\authormark{Goode, Ries, and McClernon}
\address[1]{\orgname{Sandia National Laboratories}, \orgaddress{\country{United States}}}
\corres{*Katherine Goode, Sandia National Laboratories, Albuquerque, NM. \email{kjgoode@sandia.gov}}

% Abstract and keywords 
\abstract[Abstract]{The 2022 National Defense Strategy of the United States listed climate change as a serious threat to national security. Climate intervention methods, such as stratospheric aerosol injection, have been proposed as mitigation strategies, but the downstream effects of such actions on a complex climate system are not well understood. The development of algorithmic techniques for quantifying relationships between source and impact variables related to a climate event (i.e., a climate pathway) would help inform policy decisions. Data-driven deep learning models have become powerful tools for modeling highly nonlinear relationships and may provide a route to characterize climate variable relationships. In this paper, we explore the use of an echo state network (ESN) for characterizing climate pathways. ESNs are a computationally efficient neural network variation designed for temporal data, and recent work proposes ESNs as a useful tool for forecasting spatio-temporal climate data. Like other neural networks, ESNs are non-interpretable black-box models, which poses a hurdle for understanding variable relationships. We address this issue by developing feature importance methods for ESNs in the context of spatio-temporal data to quantify variable relationships captured by the model. We conduct a simulation study to assess and compare the feature importance techniques, and we demonstrate the approach on reanalysis climate data. In the climate application, we select a time period that includes the 1991 volcanic eruption of Mount Pinatubo. This event was a significant stratospheric aerosol injection, which we use as a proxy for an artificial stratospheric aerosol injection. Using the proposed approach, we are able to characterize relationships between pathway variables associated with this event.}


\maketitle

\section{Introduction}

\begin{figure*}[h]
\centering
\includegraphics[width=\textwidth]{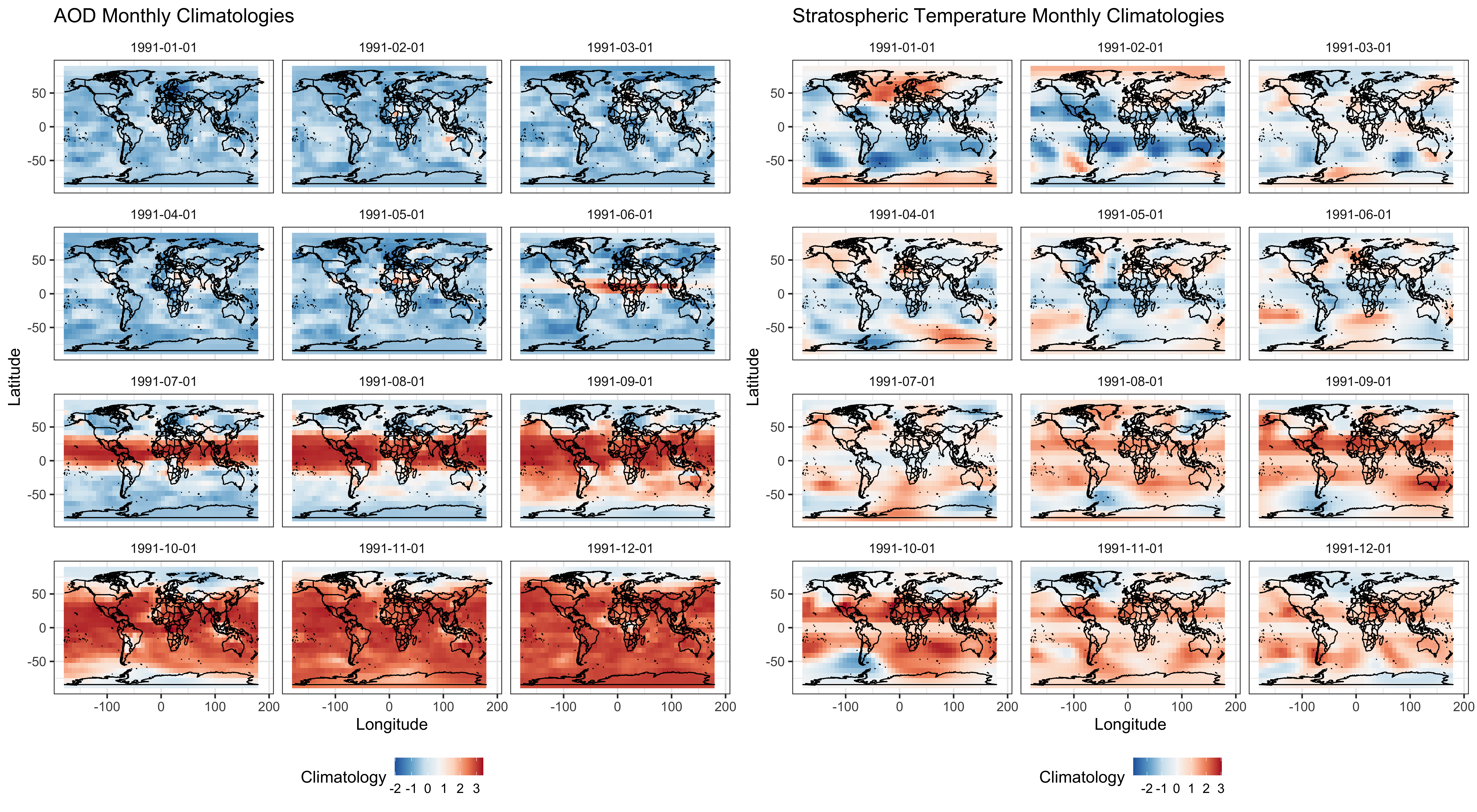}
\caption{MERRA-2 monthly climatologies in 1991 showing the effect of the Mount Pinatubo eruption in June on AOD and stratospheric temperature.}
\label{fig:merra2heatmap}
\end{figure*}

Climate change poses a serious threat to national security as acknowledged in 2022 by National Defense Strategy of the United States \citep{nds2022}. This threat is made more complicated by the possibility of artificial climate modifications. For example, strategies such as stratospheric aerosol injections, marine cloud brightening, and cirrus cloud thinning have been proposed for solar climate interventions \citep{eastham2021}. Weather modification strategies have already been implemented on regional scales such as the cloud-seeding array project `Sky River' in China, which was intended to control rainfall over the Tibetan Plateau \citep{watts2020}. While these modifications are meant as mitigation strategies for climate change, the downstream effects of such actions on a complex climate system are not well understood. The development of algorithmic methods for quantifying (i.e., characterizing) the relationships between a climate event source and its impacts would assist policy makers with high consequence decisions.

As a proxy for an artificial stratospheric aerosol injection climate event, we consider the 1991 volcanic eruption of Mount Pinatubo in the Philippines. This event has been frequently studied by climate scientists, so the relationships between the source and climate impacts are well understood. For example, the eruption released a massive injection of sulfur dioxide (SO$_2$; 18-19 Tg) into the atmosphere \citep{guo2004}, which led to increases in aerosol optical depth (AOD; a vertically integrated measure of aerosols in the air going from surface to stratosphere) \citep{sato1993, guo2004}. Ultimately, the increase in AOD resulted in stratospheric temperatures at pressure levels of 30 to 50 mb rising between 2.5 to 3.5 degrees centigrade compared to the 20-year means \citep{labitzke1992}. Figure \ref{fig:merra2heatmap} shows heatmaps of monthly climatologies (i.e., deviations from monthly spatial averages; see Equation \eqref{eq:climatology}) of AOD and stratospheric temperature in 1991 generated using Modern-Era Retrospective Analysis for Research and Applications, Version 2 (MERRA-2) data \citep{gelaro2017}. The eruption of Mount Pinatubo occurred in June of 1991, and these visualizations show the clear effects of the eruptions: above average AOD and stratospheric temperature values occurring in July through December. 

In the Mount Pinatubo example, SO$_2$ is the source variable in the climate pathway, and stratospheric temperature is the impact variable. AOD acts as an intermediate variable in the pathway. Our objective is to quantify relationships between climate pathway variables such as these. In this paper, we explore the use of echo state networks (ESNs) \citep{jaeger2001,lukosevicius:2009}, a machine learning algorithm, for the task of characterizing climate pathways. 

ESNs are known for providing good predictions with chaotic systems \citep{alao2021}, and recent work demonstrates the abilities of ESNs with long-lead forecasts on spatio-temporal climate data \citep{mcdermott2017, mcdermott2019}. ESNs are also computationally efficient models in comparison to recurrent neural networks \citep{arrieta2022}, their sibling machine learning model for temporal data, and other current statistical methods for spatio-temporal forecasting \citep{mcdermott2019}. The efficiency of ESNs is due to many parameters in the model being randomly sampled from distributions instead of estimated using gradient back-propagation as is done with recurrent neural networks. The predictive performance and computational efficiency of ESNs make them appealing models for working with large climate datasets. However, there is a clear obstacle to overcome in order to use ESNs to characterize climate pathways: lack of model \emph{interpretability}.

While there have been different definitions proposed in the literature for what makes a model \emph{interpretable} \citep{doshi2017,lipton2018,rudin2019,murdoch2019}, in this paper, we define \emph{interpretable} as follows.
\begin{definition}[Interpretable]
A model is \emph{interpretable} if it is possible to assign meaning to the model's parameters in the context of the application, which provides insight into how the model inputs relate to the model outputs.
\end{definition}
As an example, consider a linear model: $$\hat{y}=\hat{\beta_0}+\hat{\beta_1}x_1.$$ We can interpret the coefficient $\hat{\beta}_1$ as the amount the response variable $\hat{y}$ increases for a one unit increase in the predictor variable $x_1$. With an ESN model, it is not possible to assign meaning to the model parameters due to the complicated non-linear transformation applied to the input variables. Along with ESNs, many machine learning models including neural networks and random forests are classified as \emph{black-box} models due to their complex algorithms that result in this lack of interpretability. Regardless, black-box machine learning models continue to be implemented in practice due to successful demonstrations of their predictive capabilities and their data driven approach to extracting patterns in complicated applications. An approach to remedy the lack of interpretability is through \emph{explainability}.

The research area of \emph{explainable machine learning} has grown rapidly since 2015 \citep{molnar2021}. The objective of explainable machine learning is to understand how black-box models make predictions. This objective is of particular importance in high-consequence application spaces such as the medical sciences, forensics science, and national security. Since it is not possible to directly interpret black-box model parameters, many explainability approaches apply post-hoc techniques to infer how the model inputs relate to model outputs. That is, we say a model is \emph{explainable} if it satisfies the following definition.
\begin{definition}[Explainable]
A model is \emph{explainable} if it is possible to implement post hoc investigations on a trained model that infer how the model inputs relate to the model outputs.
\end{definition}
One explainability technique is the computation of feature importance (FI), which aims to quantify the effect of an input variable on a model's predictions. Various techniques have been proposed for computing FI. One example is permutation feature importance (PFI). The concept of PFI was originally introduced by \citet{breiman2001} as a FI technique for random forests and was later generalized to be model-agnostic \citep{fisher2019}. The idea with PFI is to randomly permute an input variable, while the other variables remain fixed at their observed values, and quantify how the model prediction performance is affected. Inputs that lead to the largest decrease in model performance are considered the most "important".

In this paper, we approach explainability for an ESN trained on spatio-temporal data by developing two FI techniques. Both methods approach the computation of FI by adjusting an input and quantifying how the model performance is affected similar to PFI. In fact, the first method adapts the concept of PFI to work with spatio-temporal data. PFI has been previously adapted to work with temporal data \citep{sood2021}, but to our knowledge, no work has adapted the technique for spatio-temporal. We will refer to this approach as \emph{spatio-temporal permutation feature importance (stPFI)}. Instead of permuting the values, our second approach sets the values of an input variable to zero. This essentially "turns off" the input. A similar idea was proposed as a FI technique for ESNs in \citet{arrieta2022} under the name of \emph{pixel absence effect}. In this paper, we extend the methodology to work with spatio-temporal data, and we refer to this method as \emph{spatio-temporal zeroed feature importance (stZFI)}. We compare these approaches on simulated spatio-temporal datasets and demonstrate how the proposed methods characterize climate pathways with the Mount Pinatubo example.

The remainder of the paper is organized as follows. Section \ref{sec:methods} provides the details of a single layer ESN in the context of a climate pathway scenario with spatio-temporal data and introduces the proposed techniques for computing spatio-temporal FI on ESNs. Section \ref{sec:sim} describes a simulation study implemented to compare the techniques of stPFI and stZFI for ESNs. In Section \ref{sec:merra2}, the approach is applied to characterize climate variable relationships with the Mount Pinatubo example. Finally, Section \ref{sec:conclusion} describes our conclusions and avenues for future research.

\section{Methodology} \label{sec:methods}

We consider the scenario where a spatio-temporal process that contains a known (or presumed) impact of a climate event,
\begin{align}
{\bf Z}_{Y,t} = \left(Z_{Y,t}({\bf s}_1),Z_{Y,t}({\bf s}_2),...,Z_{Y,t}({\bf s}_N)\right)',
\end{align}
is observed at a discrete set of spatial locations $\{\textbf{s}_i\in\mathcal{D}\subset\mathbb{R}^2;i=1,...,N\}$ over times $t=1,...,T$. We are interested in quantifying the relationship between this impacted spatio-temporal variable and source and/or intermediate variables pathway variables that are also observed as spatio-temporal processes:
\begin{align}
{\bf Z}_{k,t} = \left(Z_{k,t}({\bf s}_1),Z_{k,t}({\bf s}_2),...,Z_{k,t}({\bf s}_N)\right)',
\end{align} 
$k=1,...,K$. We assume that these $K$ processes are observed at the same locations and times as $\textbf{Z}_{Y,t}$, but it is possible for this assumption to be relaxed. 

We investigate the relationship between ${\bf Z}_{Y,t}$ and ${\bf Z}_{1,t},...,{\bf Z}_{K,t}$ with a two-step process:

\begin{enumerate}
\item First, we \emph{model the variable relationships} by training an ESN on times $t=1,...,T$ to forecast ${\bf Z}_{Y,t}$ using ${\bf Z}_{1,t-\tau},...,{\bf Z}_{K-1,t-\tau}$ as inputs to the model, where $\tau\in\mathbb{N}$ is the forecast lead time.
\item Next, we \emph{quantify the relationships} between the input variables and the forecasted variables using FI. For the set of forecasts at time $t$, we compute the importance of input variables at time $t$ over a block of times $\{(t-\tau), (t-\tau)-1,...,(t-\tau)-b+1\}$, where $b\in\mathbb{N}$ is the number of times in the block.
\end{enumerate}

In the rest of this section, we provide the details under this scenario of a single layer ESN and the two proposed methods for computing spatio-temporal FI. Note that much of the notation used to define the ESN is borrowed from or influenced by \citet{mcdermott2019}. All methods are implemented using R for this paper \citep{r2023}, and the code for fitting an ESN model is adapted from code provided in \citet{wikle2019}.

\subsection{Single Layer Echo State Network} \label{sec:esn}

For each spatio-temporal process, the spatial dimensions are reduced using basis functions such that for $k=1,...,K$,
\begin{align}
{\bf Z}_{Y,t}\approx\boldsymbol{\Phi}_Y\textbf{y}_{t} \ \ \ \mbox{ and } \ \ \
{\bf Z}_{k,t}\approx\boldsymbol{\Phi}_k\textbf{x}_{k,t},
\end{align} 
where $\boldsymbol{\Phi}_Y$ is an $N\times Q$ matrix of spatial basis functions and $\boldsymbol{\Phi}_k$ is an $N\times P_k$ matrix of spatial basis functions. $\textbf{y}_t$ and $\textbf{x}_{k,t}$ are vectors of length $Q$ and $P_k$, respectively, which contain the basis expansion coefficients. $Q$ and $P_k$ are user selected and are typically chosen to be much smaller than $N$. In this paper, we use principal components for the basis functions.

We create a matrix of response variables, $\textbf{Y}$, to have $Q$ rows and $T$ columns, where column $t$ contains the vector of basis functions $\textbf{y}_t$. Let $\textbf{y}_t$ represent column $t$ of $\textbf{Y}$. Then let $\textbf{X}$ be a matrix of predictor variables with $P$ rows and $T$ columns, and let $\textbf{x}_t$ represent column $t$ in $\textbf{X}$ such that $\textbf{x}_t=[\textbf{x}'_{1,t},...,\textbf{x}'_{K,t}]'$. Note that $P=\sum_{k=1}^{K}P_k$.

A single layer ESN consists of two levels:
\begin{align}
\text{\emph{Output stage:}} &\quad \textbf{y}_{t} = \mathbf{V} \mathbf{h}_t + \boldsymbol{\epsilon}_{t} \label{eq:output}\\
\text{\emph{Hidden stage:}} &\quad \mathbf{h}_t = g_h \left(\frac{\nu}{|\lambda_w|} \mathbf{W} \mathbf{h}_{t-1} + \mathbf{U} \mathbf{\tilde{x}}_{t-\tau}\right). \label{eq:hidden}
\end{align}
The input variables enter the model in the hidden stage through $\tilde{\mathbf{x}}_{t-\tau}$, which is referred to as the \emph{embedding vector} and is defined as
\begin{align}
\tilde{\mathbf{x}}_{t-\tau}=\left[\textbf{x}'_{t-\tau},\textbf{x}'_{t-\tau-\tau^*},...,\mathbf{x}'_{t-\tau-m\tau^*}\right]'.
\end{align} 
$\tau^*$ and $m$ are the \emph{embedding vector lag} and \emph{length}, respectively, which are pre-specified to determine the number of lagged inputs that are "emphasized" when computing each hidden stage. The original formulations of ESNs did not include embedding vectors (only $\textbf{x}_t$) \citep{jaeger2001,lukosevicius:2009}, but we elect to include it in our formulation since \citet{mcdermott2019} found that an embedding vector improved spatio-temporal forecasting. Additionally, in our analyses, we always use lagged inputs, so we write the ESN model with $\tilde{\textbf{x}}_{t-\tau}$, but other ESN model formulations \citep{jaeger2001,lukosevicius:2009,mcdermott2019} specify the embedding vector in the hidden stage to occur at time $t$ (i.e., $\tilde{\textbf{x}}_{t}$).

As \citet{mcdermott2019} point out, the hidden stage acts as "nonlinear stochastic transformation of the input vectors". The parameter matrices of $\textbf{W}$ and $\textbf{U}$ are referred to as \emph{reservoir weight matrices} with dimensions of $n_h\times n_h$ and $n_h\times P(m+1)$, respectively, where $n_h$ is the number of hidden units selected to include in the model. As a result, $\textbf{h}_t$ is a vector of length $n_h$ containing the \emph{hidden units}.

The elements of $\textbf{W}$ and $\textbf{U}$ are randomly sampled from distributions as follows:
\begin{align}
\textbf{W}[h,c_w] &=\gamma_{h,c_w}^w\mbox{Unif}(-a_w,a_w)+(1-\gamma_{h,c_w}^w)\delta_0,\\
\textbf{U}[h,c_u] &=\gamma_{h,c_u}^u\mbox{Unif}(-a_u,a_u)+(1-\gamma_{h,c_u}^u)\delta_0,
\end{align}
where $\textbf{W}[h,c_w]$ represents the element row $h$ and column $c_w$ of $\textbf{W}$, and similarly, $\textbf{U}[h,c_u]$ represents the element in row $h$ and column $c_u$ of $\textbf{U}$. $\gamma_{h,c_w}^w \sim Bern(\pi_w)$, $\gamma_{h,c_u}^u \sim Bern(\pi_u)$, and $\delta_0$ is a Dirac function. The values of $a_w$, $a_u$, $\pi_w$, and $\pi_u$ are pre-specified and set to small values. $a_w$ and $a_u$ are selected to prevent overfitting, and $\pi_w$ and $\pi_u$ are used to create sparse matrices.

The additional elements in the hidden stage are defined as follows:
\begin{itemize}
\item $\nu\in[0,1]$ is a pre-specified scaling parameter that helps control the amount of memory in the system,
\item $\lambda_w$ is the spectral radius of $\textbf{W}$, and
\item $g_h$ is a nonlinear activation function. Our implementation of an ESN uses a hyperbolic tangent function.
\end{itemize}

In the output stage, $\textbf{V}$ is a $Q\times n_h$ parameter matrix of coefficients estimated using a ridge regression with a penalty parameter of $\lambda_r$, and $\boldsymbol{\epsilon}_{t} \sim Gaussian\left(\mathbf{0}, \sigma_{\epsilon}^2 \mathbf{I}\right)$. Note that the only parameters estimated in the model are $\mathbf{V}$ and $\sigma^2_\epsilon$. All other parameters are randomly sampled or pre-specified, which results in the computational efficiency of the ESN model. See \citet{lukosevicius2012} for an in depth discussion of practical recommendations for ESNs including specifying tuning parameters.

Note that there are multiple locations where regularization occurs in the model. The first place is in the basis decomposition, which captures the spatial trends but reduces the dimensions and removes noise. The second place is in the output stage with the penalty parameter in the ridge regression, which drives coefficients in $\textbf{V}$ towards 0 when estimated. The third place regularization occurs in is the reservoir weight matrices. Both the sparsity that is induced in the matrices and the randomness in the generation of the matrices act as regularization mechanisms. These steps all help to prevent the ESN from over-fitting the in-sample data.

It is possible to extend this ESN model by adding terms to the output stage to account for more complicated relationships between $\textbf{h}_t$ and $\textbf{y}_{t}$. For example, in this paper, we incorporate a quadratic term in the output stage to mimic the \emph{quadratic echo state network} (QESN) described in \citet{mcdermott2019}: \begin{align}
\textbf{y}_{t} = \mathbf{V}_1 \mathbf{h}_t + \mathbf{V}_2 \mathbf{h}_t^2 + \boldsymbol{\epsilon}_{t}.
\end{align}

\subsection{ESN Feature Importance} \label{sec:fi}

\begin{figure*}
\centering
\includegraphics[width=0.95\textwidth]{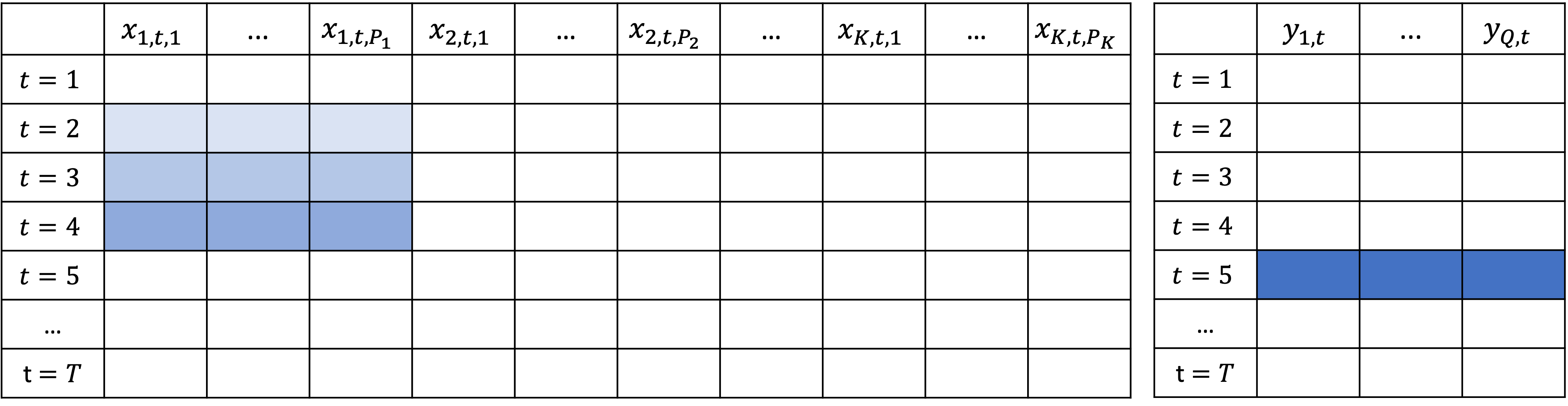}
\caption{Schematic showing an example of values in $\textbf{X}'$ (left) and $\textbf{Y}'$ (right) associated with the computation of FI $\mathcal{I}_{4,5}^{(1,3)}$.}
\label{fig:fi-demo}
\end{figure*}

Both methods that we develop for computing spatio-temporal FI for ESN models quantify "importance" through a similar concept: Determine how much model performance is affected after "adjusting" inputs at times(s) of interest in some manner. If the model performance decreases, it suggests that the input at the specified time(s) are used by the model for prediction. If the adjustment has no or little effect on the model performance, it suggests that the input at the specified time(s) are not used by the model. The larger the decrease in model performance when an input is adjusted, the larger the "importance" of the input.

With spatio-temporal data, there are various perspectives that we could consider when computing FI (e.g., blocks of time, space, or a combination). In this paper, we focus on the importance of one input spatio-temporal variable, $\textbf{Z}_{k,t}$, over a block of times, $\{t,t-1,...,t-b+1\}$, $b\in\mathbb{N}$, on the forecasts of the spatio-temporal response variable, ${\textbf{Z}}_{Y,t+\tau}$, at time $t+\tau$, averaged over locations. However, the methodology presented here could be extended to other perspectives.

Since the ESN is trained using vectors of basis expansion coefficients, $\textbf{x}_t$ and $\textbf{y}_t$, instead of the vectors $\textbf{Z}_{k,t}$ and $\textbf{Z}_{Y,t}$ on the original spatial scale, we define the FI in terms of $\textbf{x}_{t}$ and $\textbf{y}_{t}$. Recall that in our implementations, these vectors contain principal components. We will later discuss how to use back-transformations to obtain FI on the original spatial scale.

First, let $f(\textbf{x}_t, \textbf{x}_{t-1},..., \textbf{x}_{1})=\hat{\textbf{y}}_{t+\tau}$ represent the vector of forecasts from a trained ESN, $f$, at time $t+\tau$ given $\textbf{x}_t, \textbf{x}_{t-1},..., \textbf{x}_{1}$. Note that while not explicitly stated, $f$ is a function of all model parameters including $\hat{\textbf{V}}$ and $\hat{\sigma}^2_{\epsilon}$. We then let
\begin{align}
\mathcal{I}^{(k,b)}_{t,t+\tau}
\end{align} denote the FI on the trained ESN model $f$ for 
\begin{itemize}
\item spatio-temporal input variable $k$ 
\item over the block of times $\{t, t-1,..., t-b+1\}$
\item on the forecasts of the spatio-temporal response variable at time $t+\tau$. 
\end{itemize}
We compute the FI $\mathcal{I}^{(k,b)}_{t,t+\tau}$ as follows:
\begin{enumerate}
\item Obtain forecasts $f(\textbf{x}_t, \textbf{x}_{t-1},..., \textbf{x}_{1})=\hat{\textbf{y}}_{t+\tau}$ at time $t+\tau$.
\item Let $\mathcal{M}$ be a model prediction performance metric comparing observed to predicted values with the constraint that smaller values indicated better model performance (e.g., root mean squared error). Compute the performance metric on the trained model $f$ at time $t+\tau$ as: 
\begin{align}
\mathcal{M}\left(\textbf{y}_{t+\tau}, \hat{\textbf{y}}_{t+\tau}\right).
\end{align}
\item Generate \emph{adjusted} forecasts using one of the following two methods:
\begin{enumerate}
\item \emph{Permutation (stPFI)}: For replicate $r=1,2,...,R$, randomly permute the values within each vector
$\textbf{x}_{k,t}, \textbf{x}_{k,t-1},..., \textbf{x}_{k,t-b+1}$. Replace the corresponding observed values within $\textbf{x}_t, \textbf{x}_{t-1},...,\textbf{x}_{t-b+1}$ with the permuted versions. Let the versions of $\textbf{x}_t, \textbf{x}_{t-1},...,\textbf{x}_{t-b+1}$ containing the permuted values associated with variable $k$ and replicate $r$ be denoted as
\begin{align}
\textbf{x}^{(k,r)}_t, \textbf{x}^{(k,r)}_{t-1},..., \textbf{x}^{(k,r)}_{t-b+1},
\end{align}
respectively. Then obtain forecasts at time $t+\tau$ as
\begin{align}
f\left(\textbf{x}^{(k,r)}_t, \textbf{x}^{(k,r)}_{t-1},... \textbf{x}^{(k,r)}_{t-b+1}, \textbf{x}_{t-b},..., \textbf{x}_{1}\right)=\hat{\textbf{y}}^{(k,b,r)}_{t+\tau}.
\end{align}
The $R$ replications are implemented to account for variability among permutations.
\item \emph{Zeroing (stZFI)}: Replace the vectors of $\textbf{x}_{k,t}, \textbf{x}_{k,t-1},..., \textbf{x}_{k,t-b+1}$ within $\textbf{x}_t, \textbf{x}_{t-1},...,\textbf{x}_{t-b+1}$ with zeros. Let the versions of $\textbf{x}_t, \textbf{x}_{t-1},...,\textbf{x}_{t-b+1}$ containing the inserted zeros associated with variable $k$ be denoted as
\begin{align}
\textbf{x}^{(k)}_t, \textbf{x}^{(k)}_{t-1},..., \textbf{x}^{(k)}_{t-b+1},
\end{align}
respectively. Then obtain forecasts at time $t+\tau$ as 
\begin{align}
f\left(\textbf{x}^{(k)}_t, \textbf{x}^{(k)}_{t-1},... \textbf{x}^{(k)}_{t-b+1}, \textbf{x}_{t-b},...,\textbf{x}_1\right)=\hat{\textbf{y}}^{(k,b)}_{t+\tau}.
\end{align}
Note that no replications are needed to account for variability with zeroing. 
\end{enumerate}
\item Compute the prediction performance metric on the forecasts obtained by inputting the adjusted predictions into the trained model $f$. That is, with stPFI compute
\begin{align}
\mathcal{M}\left(\textbf{y}_{t+\tau}, \hat{\textbf{y}}^{(k,b,r)}_{t+\tau}\right),
\end{align}
for $r=1,...,R$, and with stZFI compute
\begin{align}
\mathcal{M}\left(\textbf{y}_{t+\tau}, \hat{\textbf{y}}^{(k,b)}_{t+\tau}\right).
\end{align}
\item Finally, either compute stPFI at time $t+\tau$ as the average change in model prediction performance when inputs $\textbf{x}_{k,t}, \textbf{x}_{k,t-1},..., \textbf{x}_{k,t-b+1}$ are permuted:
\begin{align}
\mathcal{I}^{(k,b)}_{t,t+\tau}=\left[\frac{1}{R}\sum_{r=1}^R\mathcal{M}\left(\textbf{y}_{t+\tau}, \hat{\textbf{y}}^{(k,b,r)}_{t+\tau}\right)\right] - \mathcal{M}\left(\textbf{y}_{t+\tau}, \hat{\textbf{y}}_{t+\tau}\right),
\end{align}
or stZFI at time $t+\tau$ as the change in model prediction performance when inputs $\textbf{x}_{k,t}, \textbf{x}_{k,t-1},..., \textbf{x}_{k,t-b+1}$ are set to 0:
\begin{align}
\mathcal{I}^{(k,b)}_{t,t+\tau}=\mathcal{M}\left(\textbf{y}_{t+\tau}, \hat{\textbf{y}}^{(k,b)}_{t+\tau}\right) - \mathcal{M}\left(\textbf{y}_{t+\tau}, \hat{\textbf{y}}_{t+\tau}\right).
\end{align}
\end{enumerate}

As an example, let the metric $\mathcal{M}$ used to quantify the model predictive performance be the root mean squared error (RMSE). Then FI is calculated as
\begin{multline}
 \mathcal{I}^{(k,b)}_{t,t+\tau}=\left[\frac{1}{R}\sum_{r=1}^RQ^{-1/2}\left\lVert\textbf{y}_{t+\tau}-\hat{\textbf{y}}^{(k,b,r)}_{t+\tau}\right\rVert\right]\\-Q^{-1/2}\left\lVert\textbf{y}_{t+\tau}-\hat{\textbf{y}}_{t+\tau}\right\rVert,
\end{multline}
where $\norm{\cdot}$ represents the Euclidean norm, and recall that $Q$ is the length of $\textbf{y}_t$ (i.e., the number of principal components retained for model training). In the case of zeroing, FI computed with RMSE reduces to
\begin{align}
\mathcal{I}^{(k,b)}_{t,t+\tau}=Q^{-1/2}\left(\left\lVert\textbf{y}_{t+\tau}-\hat{\textbf{y}}^{(k,b)}_{t+\tau}\right\rVert-\left\lVert\textbf{y}_{t+\tau}-\hat{\textbf{y}}_{t+\tau}\right\rVert\right).
\end{align}

While FI is defined here in terms of the principal component transformed variables, we can include a back-transformation to the spatial scale as a part of the metric $\mathcal{M}$. This will allow the interpretation of the FI values to be on a more meaningful scale. Examples of where the back-transformation is included in the performance metric function are further described and implemented in Sections \ref{sec:sim} and \ref{sec:merra2}.

Figure \ref{fig:fi-demo} provides an example schematic display of the values in the input and output matrices associated with the computation of stPFI and stZFI. The transpose of the input matrix, $\textbf{X}'$, is shown on the left where each row corresponds to $\textbf{x}'_t$. Recall that $\textbf{x}'_t=[\textbf{x}'_{1,t},...,\textbf{x}'_{K,t}]$, and let the elements of the vector $\textbf{x}'_{k,t}$ be defined as $\textbf{x}'_{k,t}=[x_{k,t,1},x_{k,t,2},...,x_{k,t,P_k}]$. The matrix on the right is the transpose of the output matrix, $\textbf{Y}'$, where each row corresponds to $\textbf{y}'_t$. Let the elements of $\textbf{y}'_t$ be defined as $\textbf{y}_t=[y_{1,t}, y_{2,t},..., y_{Q,t}]'$. This example depicts the computation of
\begin{quote}
$\mathcal{I}_{4,5}^{(1,3)}$: the importance of $\textbf{x}_{1,t}$ during the block of times $\{4,3,2\}$ $\left(\mbox{i.e., } \textbf{x}_{1,4}, \textbf{x}_{1,3}, \mbox{ and } \textbf{x}_{1,2}\right)$ on the forecasts of $\textbf{y}_{t}$ at time 5 $\left(\mbox{i.e., } \textbf{y}_5\right)$.
\end{quote}
The colored cells in $\textbf{X}'$ highlight the block of values that will be permuted/zeroed when computing stPFI/stZFI. Note that if permutation is used for computing FI, then the permutation is implemented within each row. The colored cells in $\textbf{Y}'$ are the values that will be used in the computation of the performance metric to understand the effect of the adjustment of the input values.

\section{Simulation Study} \label{sec:sim}

To assess the behavior of stZFI and stPFI, we conduct a simulation study with spatio-temporal data. The goals of this simulation are to (1) assess stZFI/stPFI on features with known differing impacts on the response and (2) determine how stZFI/stPFI is affected by varying degrees of noise in the simulated data.

\subsection{Data Generating Mechanism}

The data are generated on a lattice grid region of $[0,1]\times [0,1]$ at equally spaced locations. For $k=1,2$, let ${\bf Z}_{k,t} = (Z_{k,t}({\bf s}_1),Z_{k,t}({\bf s}_2),...,Z_{k,t}({\bf s}_N))'$ denote two spatially and temporally varying covariates. These covariates are simulated according to:
\begin{align}
{\bf Z}_{k,t} &= \mu_{k,t} + \rho_{z} {\bf Z}_{k,t-1} + \boldsymbol{\eta}_{k,t},\\
\boldsymbol{\eta}_{k,t} &\sim N \left({\bf 0}_N, \Sigma(\phi_z,\sigma^2_z)\right), 
\end{align}
for $t=2,3,...,T$, where the initial state is ${\bf Z}_{k,1} \sim N \left(\mu_{{k,1}_{N}},\Sigma(\phi_z,\sigma^2_z)\right)$. The mean functions $\mu_{k,t}$ are given by:
\begin{align}
\mu_{1,t} &= \frac{1}{\sqrt{2\pi}6} e^{-\frac{(t-20)^2}{2\times 6^2}}, \label{eq:sim-mean1}\\
\mu_{2,t} &= \frac{1}{\sqrt{2\pi}6} e^{-\frac{(t-45)^2}{2\times 6^2}}, \label{eq:sim-mean2}
\end{align}
for $t=1,2,...,T$, which result in the mean values of the covariates varying over time and peaking at $t=20$ and $t=45$, respectively. The covariance function $\Sigma$ is defined with a squared exponential kernel:
\begin{align}
\Sigma \left(\phi,\sigma^2 \right) &= \sigma^2 e^{-\frac{||s_i-s_j||^2}{2\phi^2}}. \label{eq:covariance-kernel}
\end{align}
A response is simulated by:
\begin{align}
Z_{Y,t}({\bf s}_i) &= Z_{2,t}({\bf s}_i) \beta + \delta_t({\bf s}_i) + \epsilon_t({\bf s}_i),\ \label{eq:dgm}
\end{align}
where $\epsilon_t({\bf s}_i) \overset{iid}{\sim} N(0,\sigma_{\epsilon}^2),\ \forall \ t,i$ with $t=1,...,T$ and $i=1,...,N$. The spatio-temporal random effect $\delta_t({\bf s}_i)$ is generated the same way as the covariates, letting $\boldsymbol{\delta}_t = (\delta_{t}({\bf s}_1),\delta_{t}({\bf s}_2),...,\delta_{t}({\bf s}_N))'$:
\begin{align}
\boldsymbol{\delta}_{t} &= \rho_{\delta} \boldsymbol{\delta}_{t-1} + \boldsymbol{\xi}_{t},\\
\boldsymbol{\xi}_{t} &\sim N \left({\bf 0}_N, \Sigma(\phi_{\delta},\sigma^2_{\delta})\right),
\end{align}
for $t=2,3,...,T$ with initial condition $\boldsymbol{\delta}_{1} \sim N \left({\bf 0}_N,\Sigma(\phi_{\delta},\sigma^2_{\delta})\right)$ with the same covariance function, $\Sigma$, as in Equation \eqref{eq:covariance-kernel}. 

For this study, we set $\beta=1$. Notice in Equation \eqref{eq:dgm} that the first covariate ${\bf Z}_{1,t}$ has no effect on the response ${\bf Z}_{Y,t}$. Therefore, its importance should be close to zero $\forall \ t$, while importance for ${\bf Z}_{2,t}$ should change over time as its mean values change.

We assess the effect of noise in the data on FI by considering changes in the variance parameters of the covariates, random effect and white noise terms ($\sigma_z,\sigma_{\delta},\sigma_{\epsilon}$). When generating data, we set each variance parameter to either a low variability value of 0.2 or a high variability value of 4. Additionally, we adjust the block size when computing FI, where we consider block sizes of $b=1,2,3$. We also consider changes in the spatial and correlation structures (i.e., $\phi_{z},\phi_{\delta},\rho_{z},\rho_{\delta}$), but these results are presented in the supplemental material since their effect on FI is relatively minor. 

\begin{figure}[h]
\centering
\includegraphics[width=3.5in]{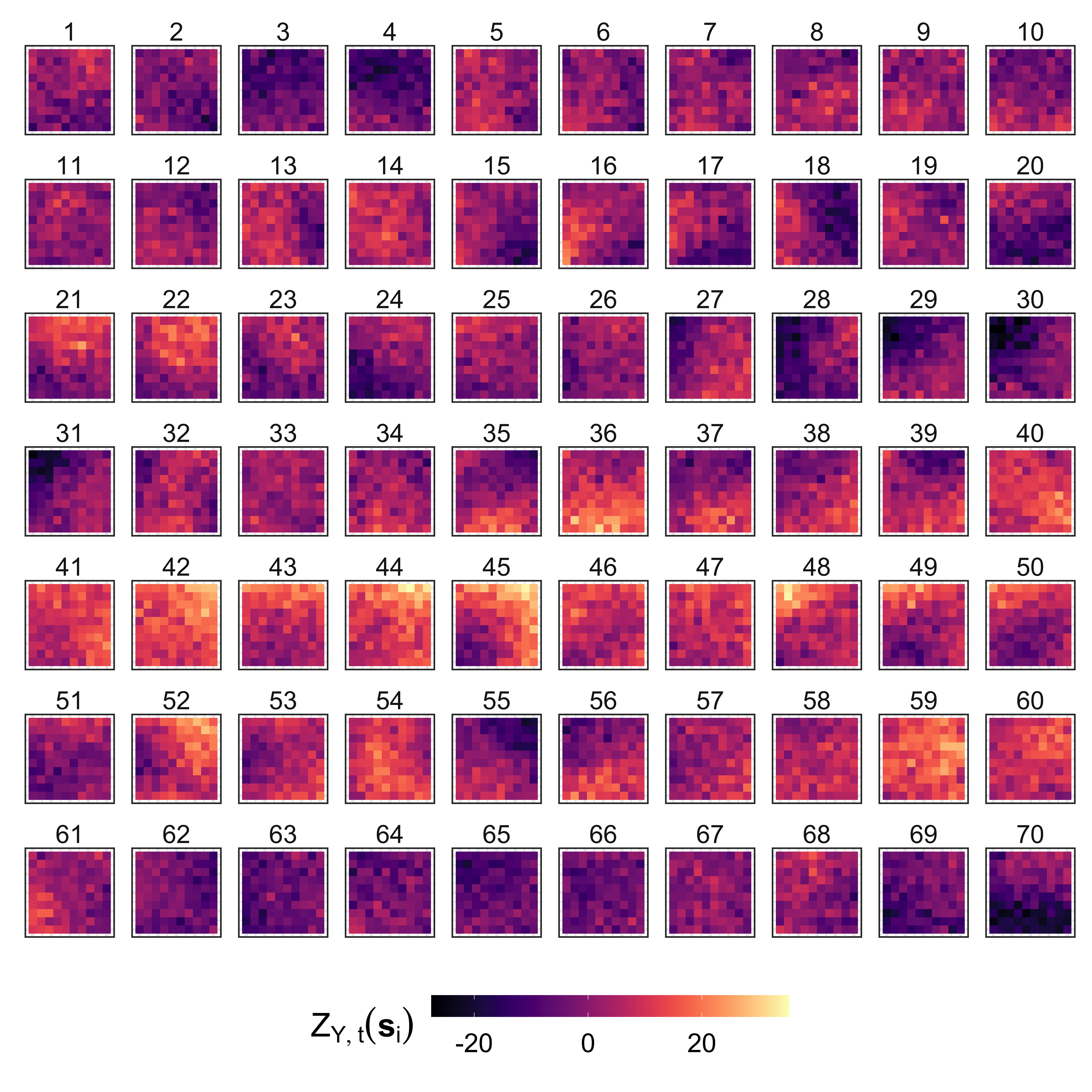}
\caption{Example of a simulated spatio-temporal data set according to Equation \eqref{eq:dgm}.}
\label{fig:simulated-data}
\end{figure}

\begin{figure}[h]
\centering
\includegraphics[width=3.5in]{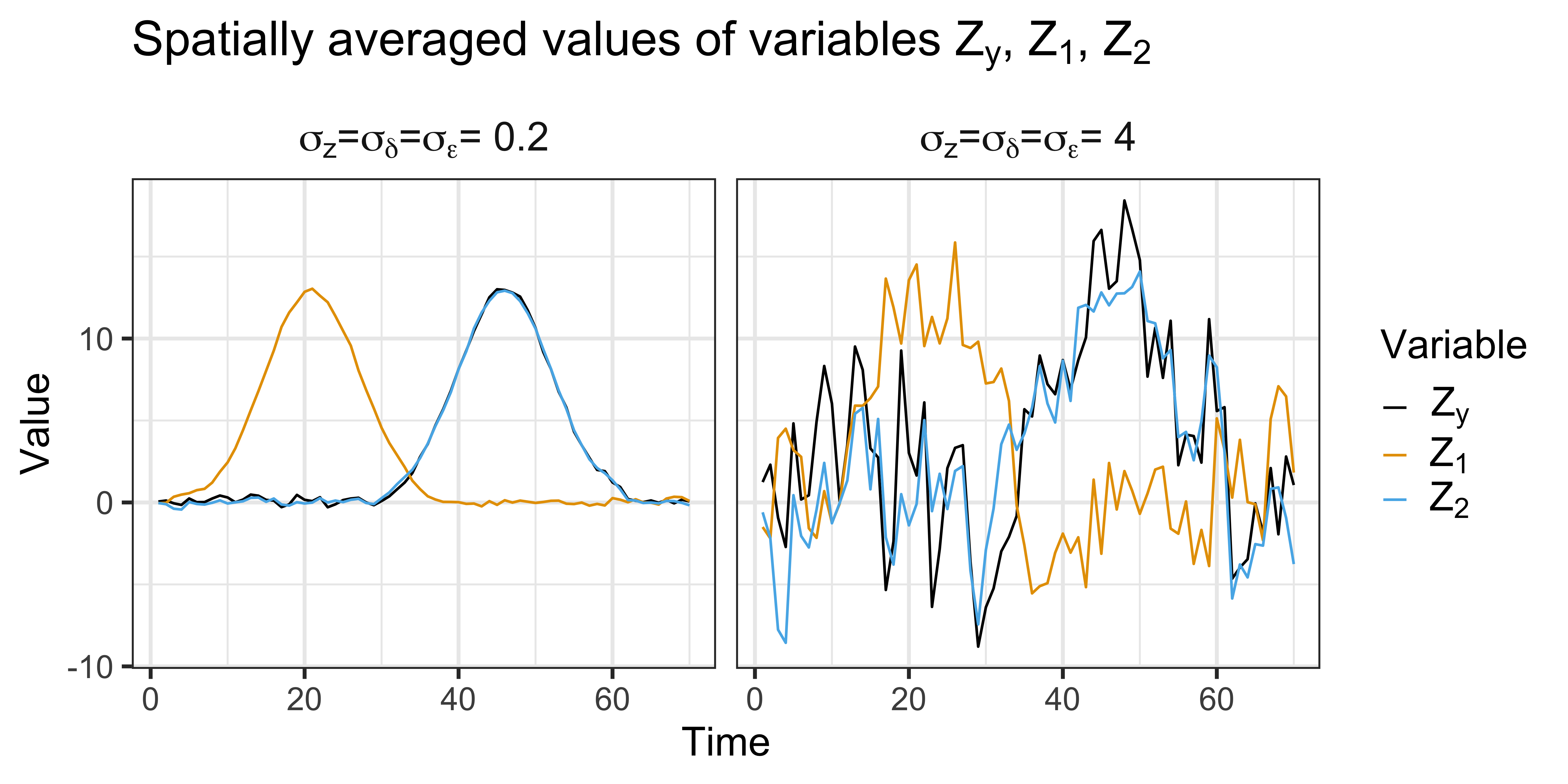}
\caption{Examples of two spatially averaged synthetic data sets: one with minimum variability and one with maximum variability. }
\label{fig:spatially-avg-simulated-data}
\end{figure}

Fifty data sets are created for each combination of parameters, and FI results are averaged over those 50. The number of time points, $T$, is set to 70, and the number of spatial locations, $N$, is set to 100, with 10 unique values in both spatial directions. Figure \ref{fig:simulated-data} shows an example of the spatio-temporal simulated response for one set of parameters. Figure \ref{fig:spatially-avg-simulated-data} shows examples of two spatially averaged data sets: one with minimum variability and one with maximum variability. 

\subsection{Models and Feature Importance}

We train an ESN to predict $\textbf{Z}_{Y,t}$ given the covariate values at a time lag of $\tau=1$: ${\bf Z}_{1,t-1}$ and ${\bf Z}_{2,t-1}$. We first standardize the response and covariates at each location by removing the sample mean across time by location and dividing by the standard deviation across time by location. Then, following the ESN set up in Section \ref{sec:esn}, we perform PCA on the standardized versions of ${\bf Z}_{1,t}$, ${\bf Z}_{2,t}$, and ${\bf Z}_{Y,t}, \ \forall \ t$. For each time $t$, the first five principal components from ${\bf Z}_{Y,t}$ make up the output vector ${\bf y}_t$, as in Equation \eqref{eq:output}. The vector of inputs at time $t$, ${\bf x}_t$, is constructed using the first five principal components from ${\bf Z}_{1,t}$ concatenated with the first five principal components from ${\bf Z}_{2,t}$, which is then used to construct the embedding vector $\tilde{{\bf x}}_t$ in Equation \eqref{eq:hidden}.

The tuning parameters for the ESN are set to $\tau^*=1,m=1,a_w=0.1,a_u=0.1,\pi_w=0.1,\pi_u=0.1,\nu=0.35$, $\lambda_r=0.1$, and $n_h=50$, and the ESN is trained using all times ($t=1,...,70$). Since $\tau=1$, $\tau^*=1$, and $m=1$, we are able to obtain forecasts for times $t=3,...,70$.

For each model, both stPFI and stZFI are computed for the two covariates (i.e., $k=1,2$) and the block sizes of $b=1,2,3$. The number of replications for stPFI, $R$, is set to 10. For the performance metric, $\mathcal{M}$, the predicted values of $\hat{\textbf{y}}_t$ are first back-transformed to the standardized spatial scale and then RMSE is computed. That is, $$\mathcal{M}\left(\textbf{y}_t,\hat{\textbf{y}}_t\right)=100^{-1/2} \left\lVert\textbf{Z}_{Y,t}-\boldsymbol{\Phi}_Y\hat{\textbf{y}}_t\right\rVert.$$

\subsection{Simulation Results}

\begin{figure*}[hbtp]
\centering
\includegraphics[width=0.92\textwidth]{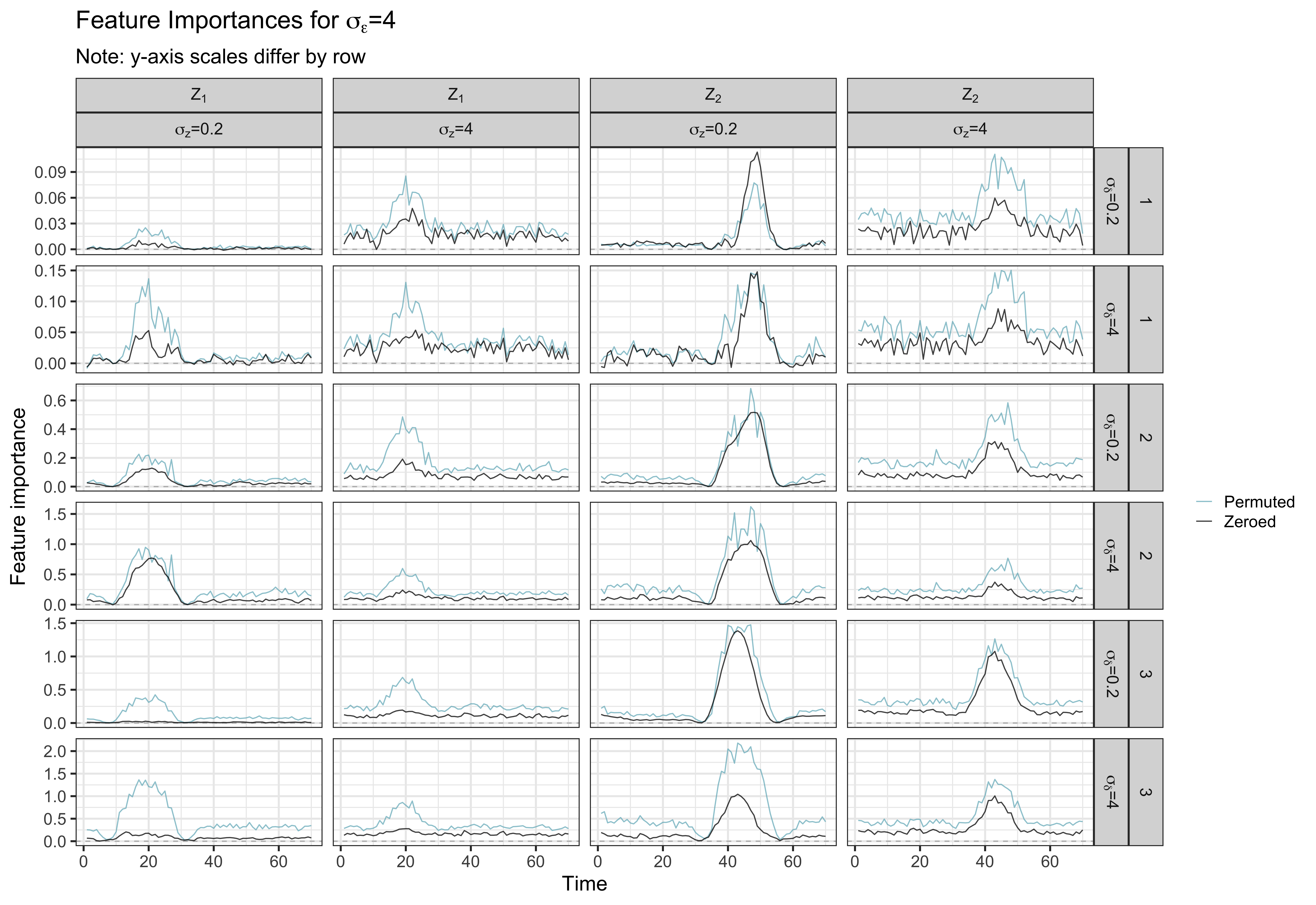}
\caption{Results from simulation study comparing stPFI and stZFI, with high level of white noise ($\sigma_{\epsilon}=4$). First two columns show FI for ${\bf Z}_1$, second two show FI for ${\bf Z}_2$. Varying combinations of $\sigma_{\delta}$ and block size are given in the rows. True data generating mechanism is given by Equation \eqref{eq:dgm}.}
\label{fig:stZFI-stPFI-comparison}
\vspace{0.5cm}
\centering
\includegraphics[width=0.92\textwidth]{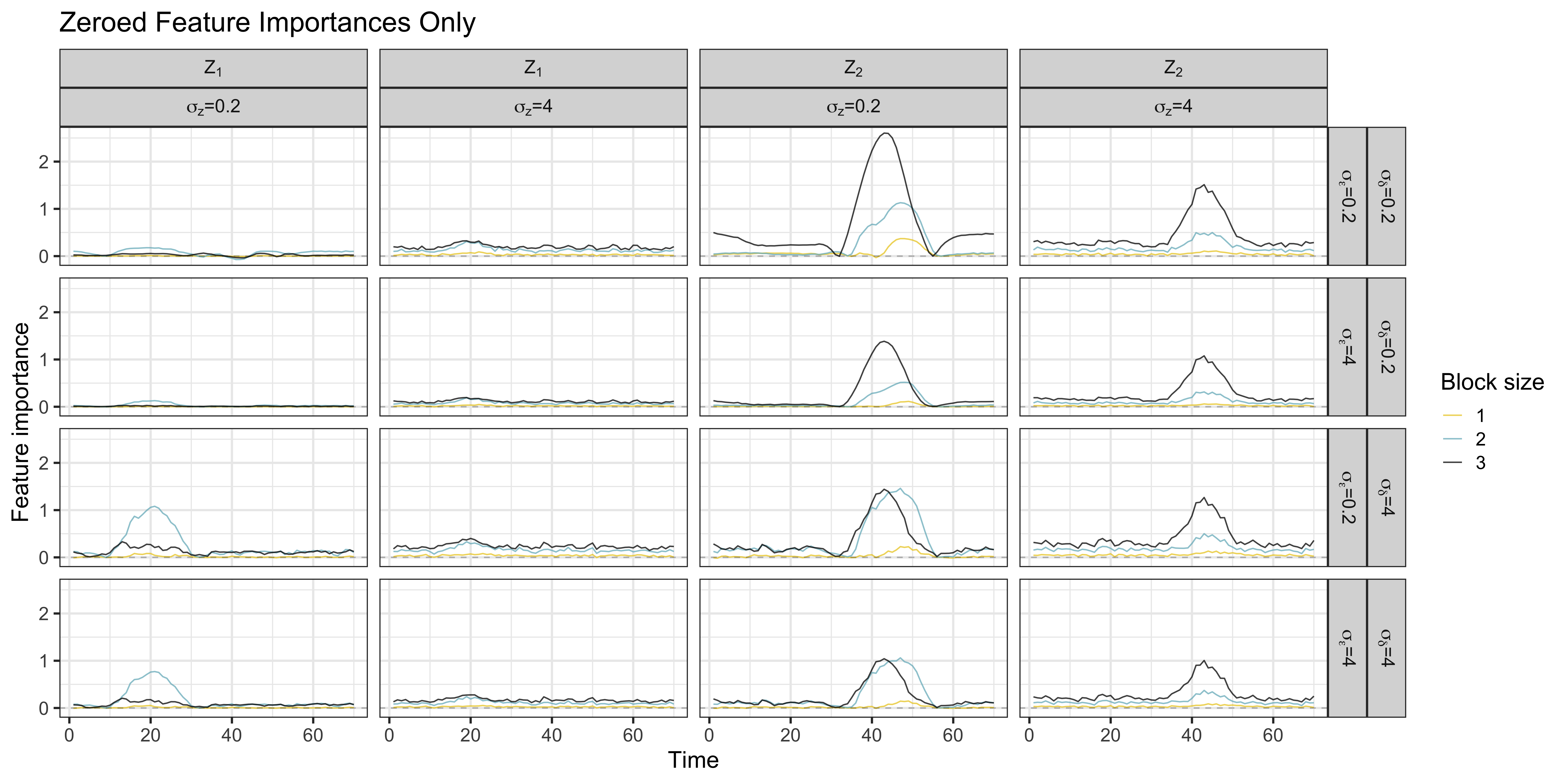}
\caption{Results from simulation study comparing number of blocks for stZFI. First two columns show FI for ${\bf Z}_1$, second two show FI for ${\bf Z}_2$. Varying combinations of $\sigma_{\delta}$ and $\sigma_{\epsilon}$ are given in the rows. True data generating mechanism is given by Equation \eqref{eq:dgm}.}
\label{fig:stZFI-nblock}
\end{figure*}

For ease of presentation, we will refer to ${\bf Z}_Y$, ${\bf Z}_1$, and ${\bf Z}_2$ generically as representing the response variable, first covariate, and second covariate, respectively. Figure \ref{fig:stZFI-stPFI-comparison} compares stPFI and stZFI in the scenario with the maximum white noise ($\sigma_{\epsilon}=4$). For all values of $\sigma_{Z}$ and $\sigma_{\delta}$ and block sizes, both stPFI and stZFI pick up on the importance of ${\bf Z}_2$ at the correct times, with peak importance around its mode at $t=45$. Changes in the variability and block size affect the FI values, but the signals are clear in the mean FI. When the block size is low, there are more fluctuations in importance values that appear to be noise, especially when variability is high. The increased block size appears to remediate this issue because it helps reduce the autocorrelation information available to the ESN.

For ${\bf Z}_1$, there are cases where both stPFI and stZFI indicate that ${\bf Z}_1$ is important when ${\bf Z}_1$ has its mode near $t=20$, but this importance is spurious since ${\bf Z}_1$ has no effect on the response. This is true for different levels of variability and for different block sizes. stZFI appears to have less of an issue with this, especially at a block size of three. We include a discussion on a probable cause for the detection of this spurious relationship and possible next steps to reduce this occurrence in Section \ref{sec:conclusion}.

Figure \ref{fig:stZFI-nblock} shows a closer view of the effect of block size on stZFI. This figure highlights that not only is stZFI smoother when the block size is increased from one to three, but the signal is also more pronounced. The increase in magnitude should be expected since as the block size increases, more of the feature times are set to zero, so the difference in RMSEs should increase, at least until the autocorrelation is removed. Thus, even with significant noise, stZFI with a large enough block-size clearly captures the importance of covariate ${\bf Z}_2$ while mostly not indicating any spurious importance of ${\bf Z}_1$.

This simulation study builds confidence in the FI approach presented in Section \ref{sec:fi}. Results show that stZFI is able to correctly identify the importance of ${\bf Z}_2$ while being relatively unaffected by ${\bf Z}_1$ when the block size parameter is sufficiently large, and these results translate across varying noise levels. stPFI has similarly strong performance identifying ${\bf Z}_2$ but is more susceptible to detecting importance in a variable that has no direct impact on the response. Additional figures of simulation results are provided in the supplemental material.

\section{Climate Data Application} \label{sec:merra2}

\begin{figure*}
\centering
\includegraphics[width=\textwidth]{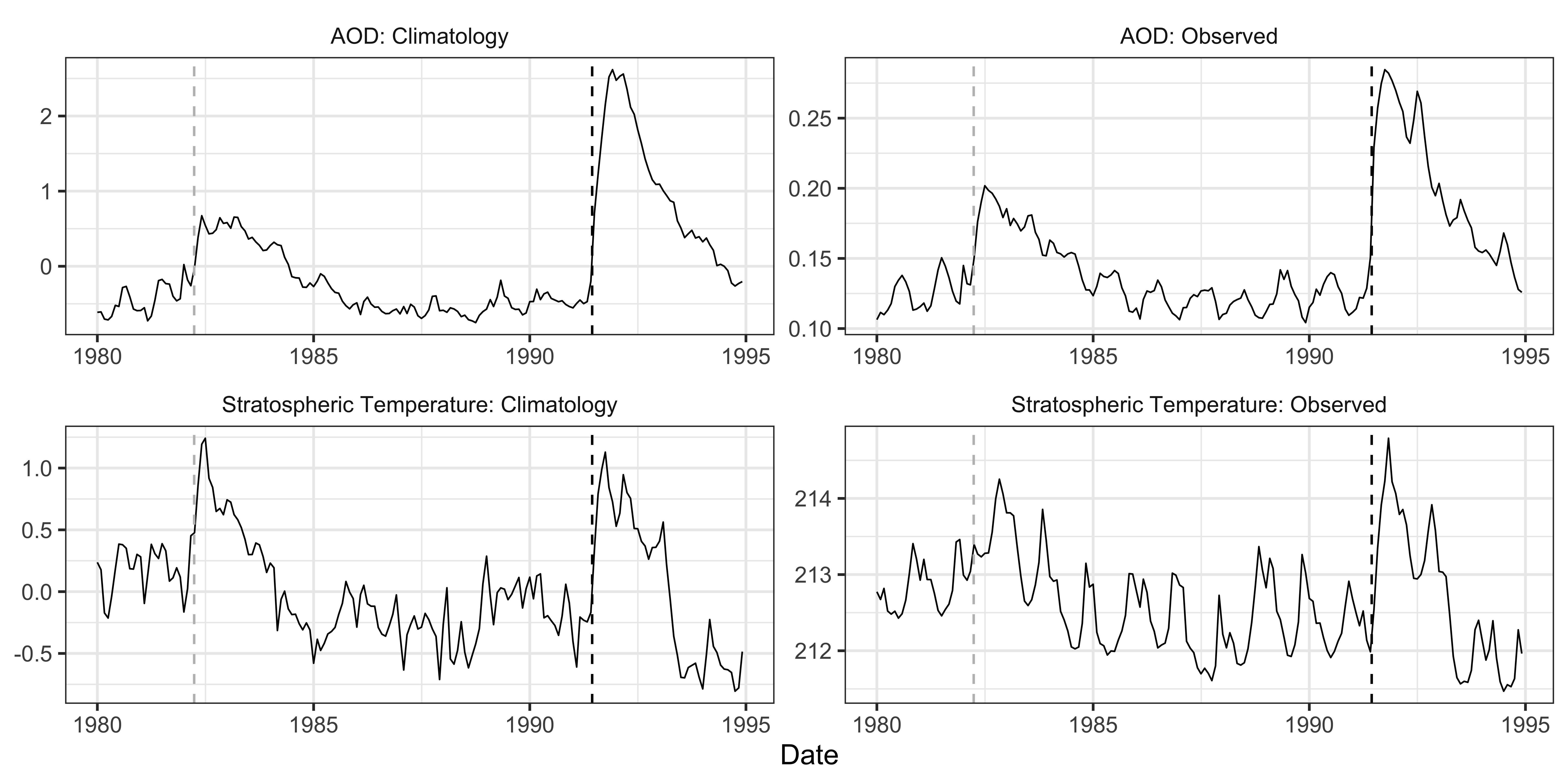}
\caption{MERRA-2 global weighted averages of stratospheric temperature and AOD. Weighting is described in Equation \eqref{eq:weighting}. Climatologies remove average monthly effect by spatial location (see Equation \eqref{eq:climatology}). The gray vertical dashed line indicates the eruptio of El Chich\'{o}n, and the black vertical dashed line indicates the eruption of Mount Pinatubo.}
\label{fig:merra2}
\end{figure*}

On June 12, 1991, Mount Pinatubo erupted in the Philippines, having a profound impact on the climate. The eruption released 18-19 Tg of SO$_2$ into the atmosphere \citep{guo2004}. The massive injection of aerosols into the atmosphere led to increases in AOD \citep{sato1993, guo2004}, which in turn led to changes in stratospheric temperatures and surface temperatures \citep{sato1993, liepert2002} (and references therein).

When aerosols enter the atmosphere they can either scatter sunlight, which leads to atmospheric cooling, or absorb the sunlight, which leads to warming \citep{liepert2002,myhre2013}. Due to the relationship between atmospheric aerosols and temperatures, the artificial injection of aerosols has been discussed as a potential mitigation to the current climate change trends. This was foreshadowed by \citet{kiehl1993} in 1993 who noted that summer sulfate aerosol forcings offset greenhouse forcings in the eastern US and central Europe. However, there is great uncertainty in how such an intervention would affect the broader climate system.

We aim to explore the effects of anthropogenic forcings of aerosols into the atmosphere by using a black-box model (i.e., an ESN) to quantify their impacts on the climate system. We use the 1991 Mount Pinatubo eruption as an proxy for anthropogenic injection of aerosols. We will focus on the relationship between the climate pathway variables of AOD and stratospheric temperatures, but future work could explore the inclusion of additional pathway variables such as SO$_2$ and surface temperatures.

To analyze this question, we use the Modern-Era Retrospective Analysis for Research and Applications, Version 2 (MERRA-2) \citep{gelaro2017} for stratospheric data at 50 mb \citep{merra2temp} and vertically integrated AOD \citep{merra2aod}. Detailed information on AOD from MERRA-2 is provided in \citet{randles2017}. We consider the years of 1980-1995, which provides climate information before the eruption of Mount Pinatubo and includes a second climate event: the 1982 eruption of El Chichón, in southeast Mexico, which injected 7.5 Tg of SO$_2$ into the atmosphere \citep{krueger2008}.

The data for both AOD and stratospheric temperature are on the monthly time scale and a 24$\times$48 equally spaced latitude and longitude lattice. Figure \ref{fig:merra2} shows global weighted average of stratospheric temperatures and aerosol optical depth (AOD) over this time period for observed and monthly climatological values. The computation of climatologies will be described in Equation \eqref{eq:climatology}, and the weighting will be described in Equation \eqref{eq:weighting}. The effects of Mount Pinatubo and El Chichón are clear in both variables: increases in AOD result in increases in stratospheric temperature immediately following the eruptions. This trend is expected due to the reflection of the sun's energy \citep{labitzke1992}.

\begin{figure*}[ht!]
\centering
\includegraphics[width=\textwidth]{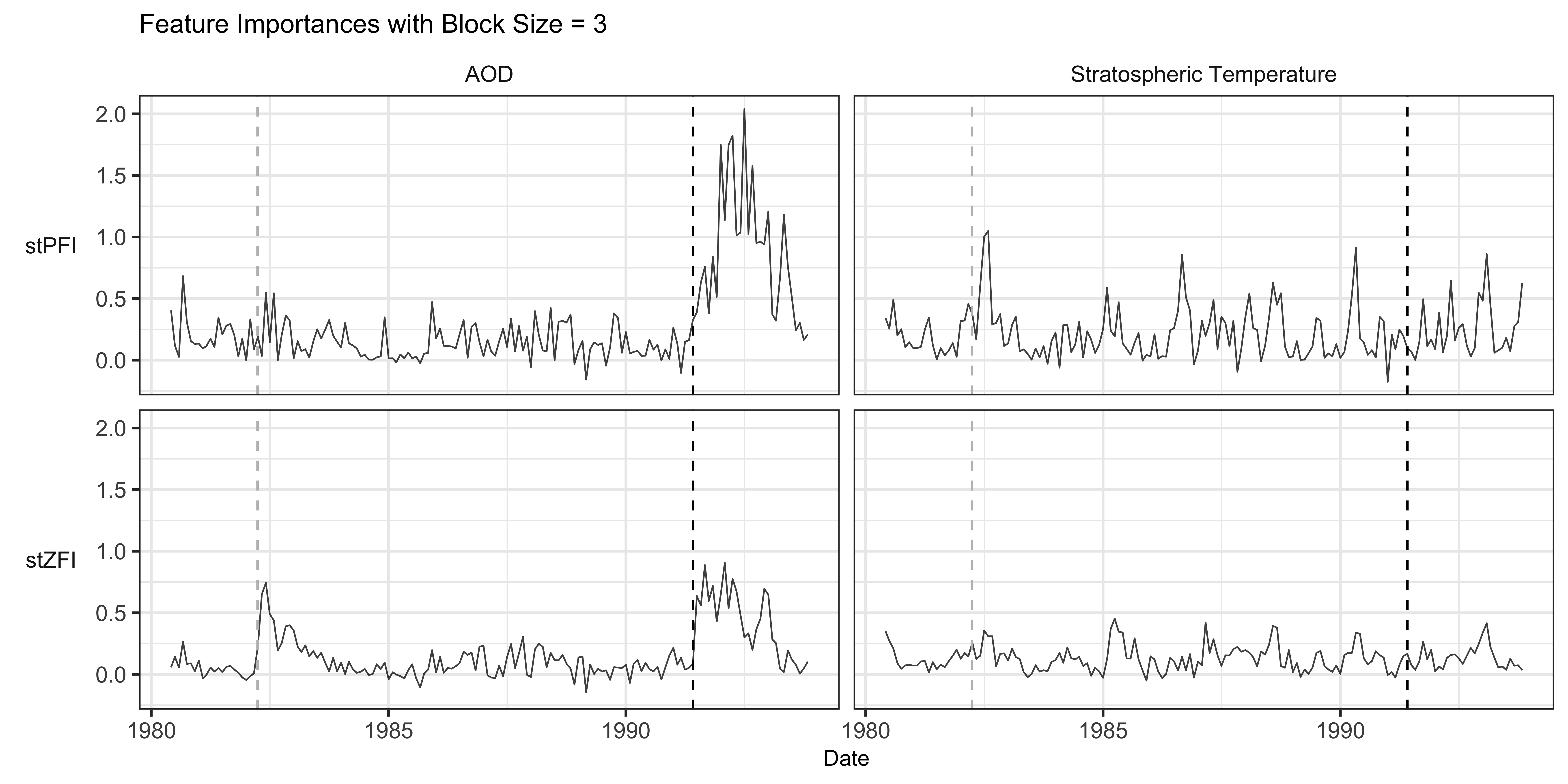}
\caption{Feature importances on MERRA-2 data with a block size of 3. The gray vertical dashed line indicates the eruptio of El Chich\'{o}n, and the black vertical dashed line indicates the eruption of Mount Pinatubo.}
\label{fig:merra2fi}
\end{figure*}

Because temperature, and to a lesser extent, AOD, exhibits strong seasonality, preprocessing of the data is done. We use monthly \emph{climatologies}, which removes the mean monthly effect and divides by the standard deviation of the monthly effect. Let ${\bf s}_i,\ i=1,2,...,N$ be unique latitude/longitude locations. In a slight change of notation, let $Z_{k,mth,yr}({\bf s}_i)$ be the raw, observed value of variable $k$ at location ${\bf s}_i$, month $mth$, and year $yr$. Let $k=1$ denote AOD and $k=2$ denote stratospheric temperature. We are interested in forecasting one month ahead (i.e. $\tau=1$) stratospheric temperatures, $Z_{Y,mth,yr}({\bf s}_i)$. The climatologies are calculated according to:

\begin{align}
Z^{clmt}_{k,mth,yr}({\bf s}_i) &= \frac{Z_{k,mth,yr}({\bf s}_i)-\bar{Z}_{k,mth,\cdot}({\bf s}_i)}{sd(Z_{k,mth,\cdot}({\bf s}_i))}, \label{eq:climatology}
\end{align}

\noindent where $\bar{Z}_{k,mth,\cdot}({\bf s}_i)$ is the average temperature at location ${\bf s}_i$ during month $mth$, for variable $k$, and $sd(Z_{k,mth,\cdot}({\bf s}_i))$ is the standard deviation of temperatures at location ${\bf s}_i$ during month $mth$. Climatologies for $Z_{Y,mth,yr}({\bf s}_i)$ are calculated in the same manner, and are equal to the climatologies of $Z_{2,mth,yr}({\bf s}_i)$ one time step ahead. 

The ESN is trained using data from 1980-1995. Here, we use the first five principal components from both stratospheric temperature and AOD as an example where the spatial dimensions are greatly reduced (1,152 locations reduced to 5 principal components), but future work could explore method performance with different numbers of principal components. The ESN embedding vector is constructed using $m=5$, $\tau=1$, $\tau^*=1$, meaning the ESN predicts stratospheric temperatures one month ahead using AOD and stratospheric temperatures from the previous five months. The tuning parameters for the ESN are the same as for the simulation study, $a_w=0.1,a_u=0.1,\pi_w=0.1,\pi_u=0.1,\nu=0.35$, $\lambda_r=0.1$, and $n_h=50$. A sensitivity analysis shows that $\lambda_r$ has the biggest effect of these hyperparameters. Hyperparameters associated with ${\bf W}$ have a slightly bigger effect than hyperparameters associated with ${\bf U}$, but they are still secondary to $\lambda_r$. FI is calculated on the training data since we are ultimately interested in variable relationships, not forecasting. However, we need to ensure the model fits the data well, so we believe the FI are meaningful. A time series blocked training/test split evaluation is provided in the supplemental material.

Since the data are on an equally spaced lattice, there are relatively more locations towards the poles compared to the equator. To mitigate the effect of poles (since they are more variable temperature-wise), we opt to use a weighted RMSE as our metric for FI. Taking the square root of the cosine of the latitude has been shown to be a good way of weighting latitudes \citep{huth2006}. Let $\hat{Z}^{clmt}_{Y,mth,yr}({\bf s}_i)$ be the model's prediction of stratospheric temperature at location ${\bf s}_i$, month $mth$, year $yr$, which is back-transformed from the principal component scale. Then the weighted RMSE is calculated by:

\begin{align}
RSE_{mth,yr}({\bf s}_i) &= \sqrt{\left( Z^{clmt}_{Y,mth,yr}({\bf s}_i)-\hat{Z}^{clmt}_{Y,mth,yr}({\bf s}_i) \right)^2}, \nonumber \\
Weighted\ RMSE_{mth,yr} &= \frac{\sum_{i=1}^{N} w_{{\bf s}_i} RSE_{mth,yr}({\bf s}_i)}{\sum_{i=1}^{N} w_{{\bf s}_i}}, \nonumber \\
w_{{\bf s}_i} &= \sqrt{cos \left( latitude({\bf s}_i) \times \frac{\pi}{180} \right)}, \label{eq:weighting}
\end{align}

\noindent where $latitude({\bf s}_i)$ returns the latitude of location ${\bf s}_i$ in degrees.

Figure \ref{fig:merra2fi} shows stPFI and stZFI on the MERRA-2 data computed with a block size of three. A figure in the supplemental material shows comparisons with different block sizes. The two vertical dashed lines show the eruptions of El Chichón and Mount Pinatubo. The effects of Mount Pinatubo are clear for AOD, as both stPFI and stZFI see a large spike in importance. This indicates the importance of AOD for making one month ahead forecasts of stratospheric temperature. Lagged stratospheric temperatures do not see as much of a change after the Pinatubo eruption, suggesting standard temperature fluctuations are not sufficient for explaining the changes in temperature. These two pieces provide evidence, but not proof, that the impact on temperature due to the volcanic eruption and its subsequent injection of aerosols can be traced through AOD.

stZFI also captures the effect from El Chichón in AOD. The importance diminishes faster than after Mount Pinatubo, but this is not surprising since Mount Pinatubo was a much bigger eruption. stPFI on the other hand, does not appear to find AOD particularly important after El Chichón, but instead stratospheric temperature has a peak of importance after this eruption. Although the effects of El Chichón and Mount Pinatubo have been well studied, this proof of concept showcases the methodology's ability to recapture known effects.

\section{Conclusions and Discussion} \label{sec:conclusion}

In this paper, we propose the use of ESNs for characterizing climate pathways (i.e., quantifying relationships between climate variables related to a climate event). We do this by modeling spatio-temporal climate pathway variables associated with a climate event using an ESN and quantifying the variable relationships using FI.

ESNs are a computationally efficient model that are able to capture patterns in complex systems, which makes them a desirable tool for applications with the complex climate system containing large quantities of data. In order to provide transparency to the black-box ESN, we develop two FI techniques (stPFI and stZFI) for spatio-temporal data that are applied to the ESN in order to quantify the variable relationships captured by the ESN. Both FI techniques approach the quantification of variable importance by adjusting (permuting or setting to zero) a block of times associated with a climate variable of interest and measure how this adjustment affects the model forecast performance at a specified time. By visualizing the resulting FIs, we depict how the importance of input variables on the forecast variable changes over time and compare the importance values to other input variables.

We demonstrate our approach on MERRA-2 reanalysis climate data that included two volcanic eruption events (El Chich\'{o}n in 1982 and Mount Pinatubo in 1991), which act as proxies of artificial stratospheric aerosol injections. We consider the relationships between the pathway variables of AOD and stratospheric temperature. The ESN FI results show that the importance of AOD on forecasting stratospheric temperature greatly increases after both eruptions, which provides support for AOD being a part of the climate pathway leading to the effects seen in stratospheric temperature. These results agree with previous climate science research indicating that the increase in AOD due to the eruptions led to an effect on the temperature, which supports the capabilities of the method.

In future work, additional variables in the Mount Pinatubo pathway such as SO$_2$ and surface temperature could be integrated into the methodology. The inclusion of additional variables could take the form of additional inputs to the model, or in some scenarios, it may be of interest to consider the joint forecasting of multiple variables (e.g., surface and stratospheric temperatures). In addition, when computing FI, it may be meaningful to consider the importance of groups of climate variables (e.g., AOD and SO$_2$). By grouping variables when computing FI, the results can be interpreted as the joint effect. This approach could be especially useful with highly correlated pathway variables.

In this paper, we compute FI as an average over locations, so that it reduces to a measure of importance over time. Another option would be to adjust the computation of FI in a manner such that a measure of importance is associated with each location (or specified regions such as latitudinal bands at a time) at a time/block of times. This approach would allow for identification of how importance not only changes over time but also over space.

Further development of FI techniques for spatio-temporal data could also include steps that better account for correlation in the data. In the simulation study, both stPFI and stZFI are able to pick up on the variable known to be related to the response ($\textbf{Z}_2$), and when stZFI is applied with larger block sizes, it is less likely to pick up on the spurious variable relationship ($\textbf{Z}_1$). However, stPFI always detects the spurious variable relations, and in many cases, stZFI incorrectly identifies this relationship. It is likely the case that these results are due to correlation in the data. It has been suggested that permutation based FI methods produce biased results in the data when correlation is present and not accounted for \citep{hooker2021}. This bias is due to the permutation leading to observations that occur outside of the observed training data, which leads to model extrapolation and inaccurate measures of variable importance. It seems reasonable that this same phenomenon could occur when `zeroing' the data but to a lesser extent. This could explain why stZFI is less affected than stPFI. \citet{hooker2021} suggest overcoming this issue by either retraining the model on the permuted data or developing a conditional FI. These ideas would be interesting to explore in the case of spatio-temporal FI, where the existence of correlation is essentially guaranteed.

In addition to different approaches to computing FI, future work could consider ESN extensions. We incorporate a single-layer ESN in our methodology in this paper, but the methodology could easily be extended to work with other variations of ESNs. Extensions to ESNs include the addition of multiple layers for capturing temporal trends on different time scales, model ensembles, and Bayesian implementations for quantifying uncertainty \citep{mcdermott2019}.

Other approaches to modeling climate data with deep learning have included convolutional neural networks (CNNs) (e.g., \citet{mamalakis2022}) and Bayesian neural networks (BNNs) (e.g., \citep{clare2022}). These approaches then used explainability techniques such as Layer-wise Relevance Propagation (LRP) \citep{bach2015} and SHapley Additive exPlanation (SHAP) values \citep{lundberg2017}. A recent work also demonstrated how LRPs can be applied to ESNs in the context of a climate application \citep{landt2022}. Future work could compare other deep learning and explainability techniques in terms of both computation time and variable relationships identified.

As the possibility of the implementation of climate mitigation strategies becomes more of a reality, the importance of the development of algorithmic tools for understanding how such actions could affect the other aspects of the climate increases. We approach this task by quantifying pathway variable relationships using FI computed on an ESN trained over a time period surrounding the event of interest. However, just as the ESN is an approximation to the workings of the climate system, FI is an approximation to the workings of the ESN. Further development of explainability techniques for spatio-temporal data that provide different perspectives on black-box models or the development of machine learning models for spatio-temporal data with interpretable parameters could lend more credibility to the use of machine learning models in such high-stakes applications.

\section*{Acknowledgements} 

The authors thank Lyndsay Shand, Gabriel Huerta, and J. Derek Tucker for their thoughtful suggestions during development stages. Additionally, we thank Gabriel Huerta for his careful read through and feedback on the paper.

\section*{Funding Statement} 

This work was supported by the Laboratory Directed Research and Development program at Sandia National Laboratories, a multimission laboratory managed and operated by National Technology and Engineering Solutions of Sandia LLC, a wholly owned subsidiary of Honeywell International Inc. for the U.S. Department of Energy's National Nuclear Security Administration under contract DE-NA0003525. This paper describes objective technical results and analysis. Any subjective views or opinions that might be expressed in the paper do not necessarily represent the views of the U.S. Department of Energy or the United States Government. SAND2023-10821O.

\section*{Supporting Information} 

Additional supporting information can be found online in the Supporting Information section at the end of this article.

%\clearpage 

\bibliographystyle{plain}
\bibliography{references}

\section*{Supplemental Material}

This document contains additional results from the simulations study and the Mount Pinatubo example.

\subsection*{Simulation Study Additional Results}

\begin{figure*}[h]
\centering
\includegraphics[width=\textwidth]{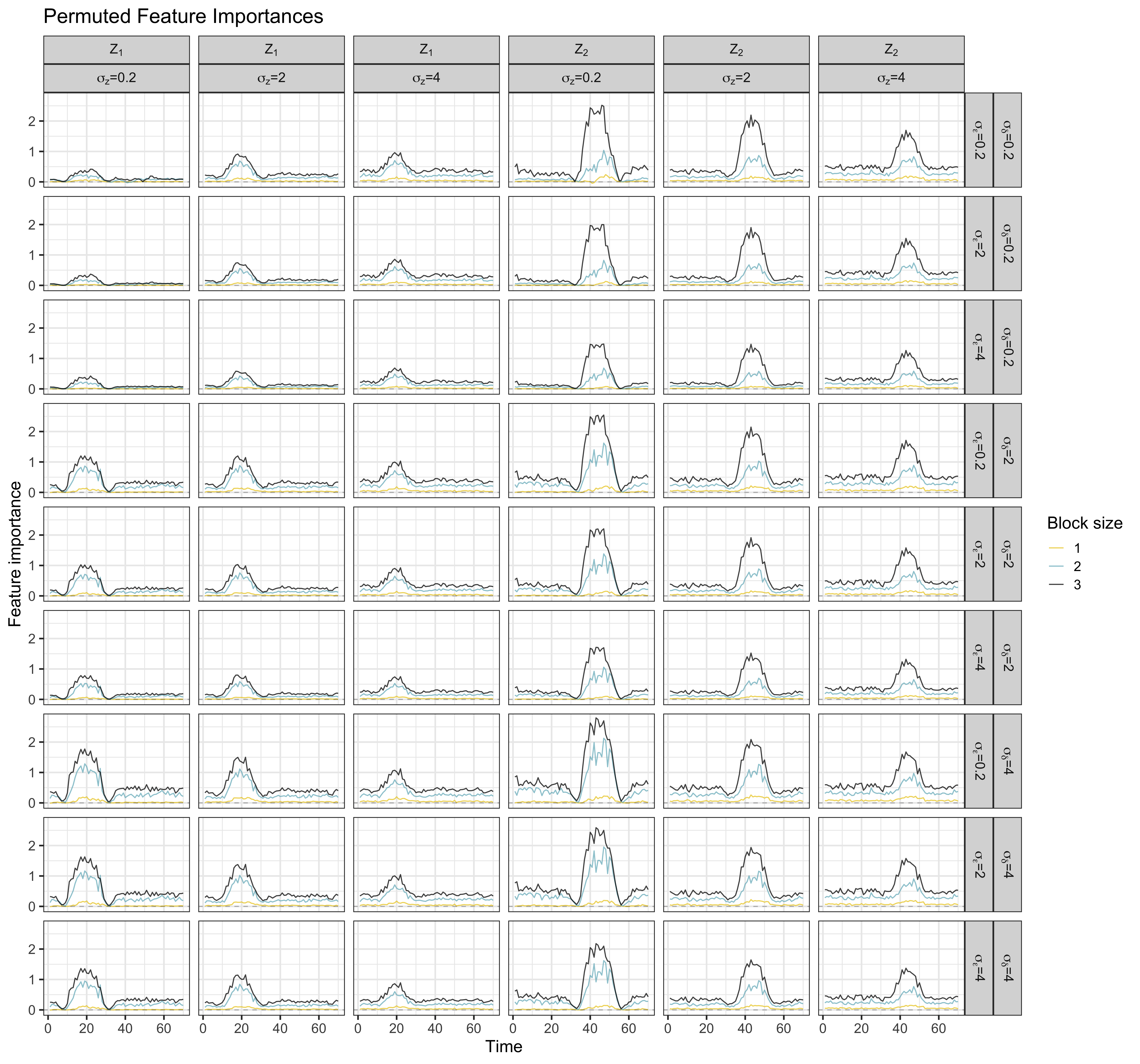}
\caption{Detailed simulation study results for stPFI. These plots explore changes to $\sigma_z,\sigma_{\delta},\sigma_{\epsilon}$ and number of blocks.}
\label{fig:pfi-supplement}
\end{figure*}

Figure \ref{fig:pfi-supplement} shows stPFI for variables ${\bf Z}_1$ and ${\bf Z}_2$, while varying $\sigma_z,\sigma_{\delta},\sigma_{\epsilon}$ and number of blocks. stPFI tends to be more pronounced with an increase in the block size, as well as smoother. stPFI does pick up on importance for ${\bf Z}_1$ even though ${\bf Z}_1$ has no direct impact on the response, and worse, this effect grows with increasing block size. An increase in the variation of $\sigma_z$ corresponds with a flattening of stPFI across the board which is not surprising since the true importance is being masked by noise. The random effect variability $\sigma_{\delta}$ and white noise variability $\sigma_{\epsilon}$ do not have a large impact on stPFI, but increases of $\sigma_z$ do reduce the FI values in magnitude and in the range of when stPFI picks up the importance. Regardless of the standard deviations being increased by an order of magnitude, stPFI still has a clear signal.

\begin{figure*}[h]
\centering
\includegraphics[width=\textwidth]{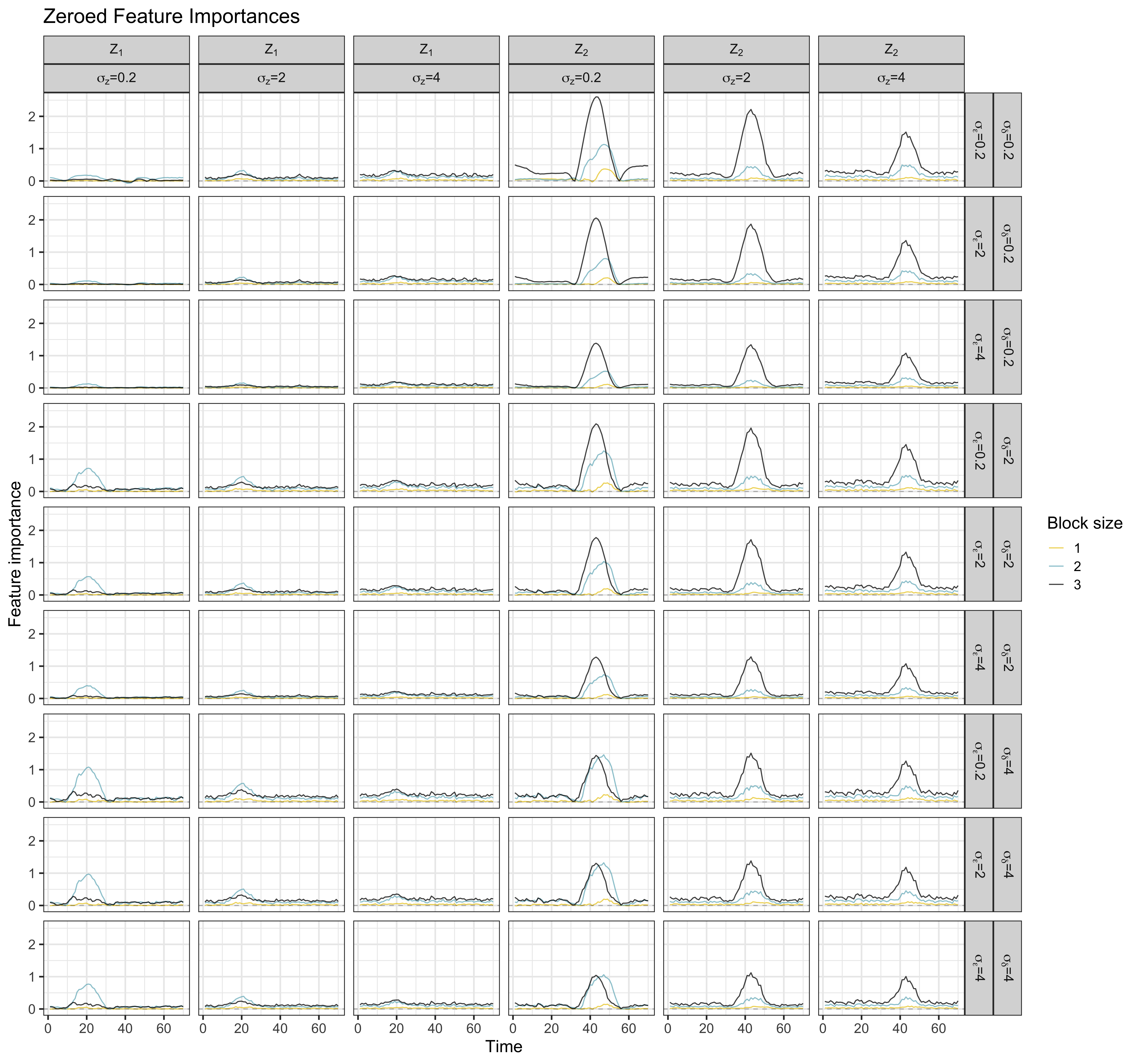}
\caption{Detailed simulation study results for stZFI. These plots explore changes to $\sigma_z,\sigma_{\delta},\sigma_{\epsilon}$ and number of blocks.}
\label{fig:zfi-supplement}
\end{figure*}

Figure \ref{fig:zfi-supplement} shows stZFI for variables ${\bf Z}_1$ and ${\bf Z}_2$, while varying $\sigma_z,\sigma_{\delta},\sigma_{\epsilon}$ and number of blocks. stZFI tends to be more pronounced with an increase in the block size, as well as more smooth. In some situations, stZFI does pick up on importance for ${\bf Z}_1$ even though ${\bf Z}_1$ has no direct impact on the response. However, this effect is much smaller in magnitude compared to the true importance of variable ${\bf Z}_2$. An increase in the variation of $\sigma_z$ corresponds with a flattening of stZFI across the board, as well as less smooth FI, which is not surprising since the true importance is being masked by noise. The random effect variability $\sigma_{\delta}$ and white noise variability $\sigma_{\epsilon}$ do not have a large impact on stZFI, but increases of $\sigma_z$ do reduce the FI values in magnitude and in the range of when stZFI picks up the importance. Regardless of the standard deviations being increased by an order of magnitude, stZFI still has a clear signal.

\begin{figure*}[h]
\centering
\includegraphics[width=\textwidth]{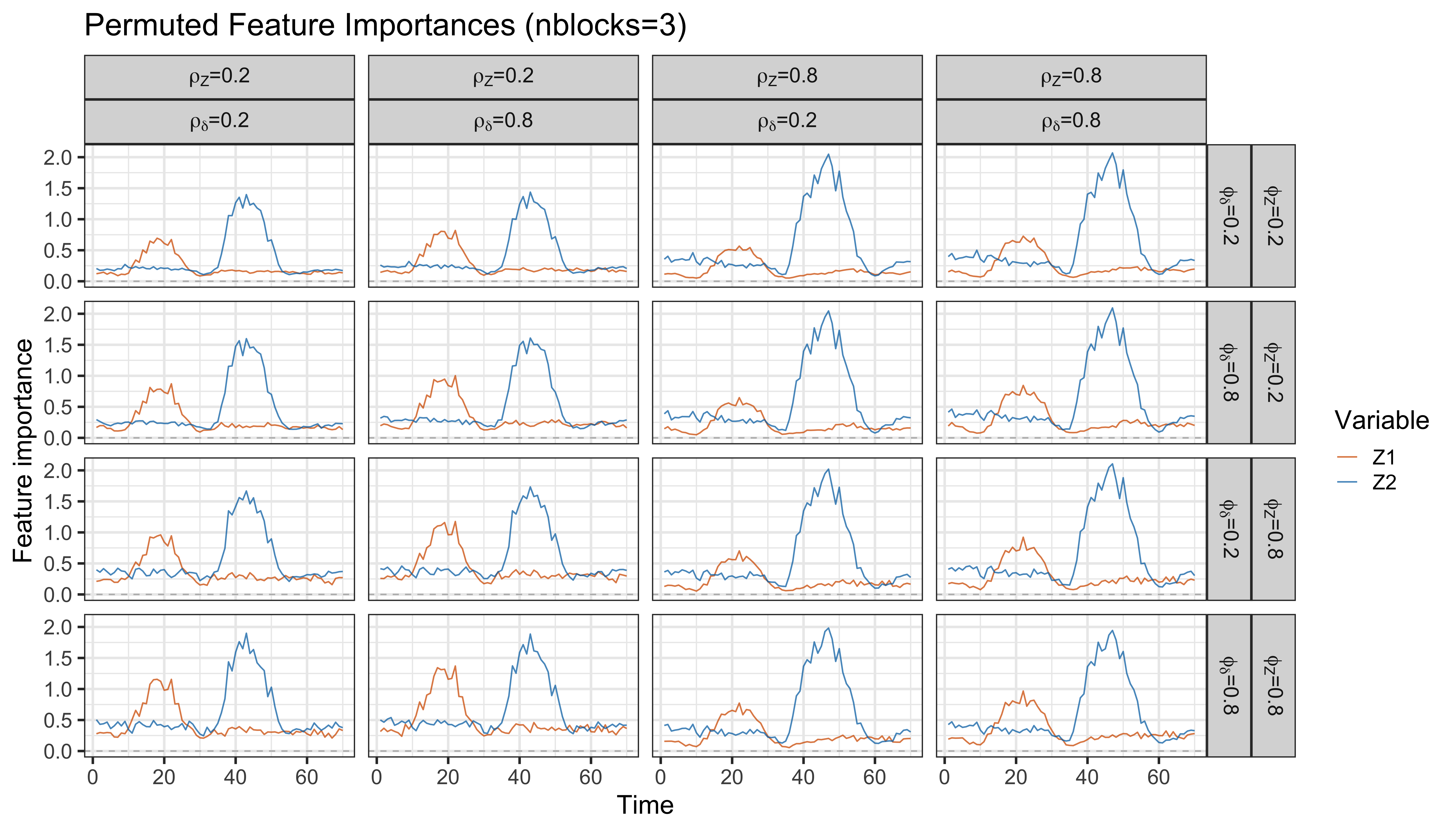}
\caption{Detailed simulation study results for stPFI and block size of 3. These plots explore changes to $\rho_z,\rho_{\delta},\phi_z,\phi_{\delta}$.}
\label{fig:pfi-supplement-rhophi}
\end{figure*}

Figure \ref{fig:pfi-supplement-rhophi} shows stPFI for variables ${\bf Z}_1$ and ${\bf Z}_2$, while varying $\rho_z,\rho_{\delta},\phi_z,\phi_{\delta}$ with a block size of 3. We restricted block size to 3 for brevity, because there did not appear to be any interaction between block size and these parameters. The main difference is when $\rho_z$ is smaller, stPFI is relatively larger for ${\bf Z}_1$ and relatively smaller for ${\bf Z}_2$ compared to when $\rho_z$ is large, meaning the performance of stPFI is better with larger $\rho_z$. Larger autocorrelation of the random effect, $\rho_{\delta}$, appears to make stPFI perform worse (relatively larger stPFI for ${\bf Z}_1$ and relatively smaller stPFI for ${\bf Z}_2$). The spatial range parameters, $\phi_z,\phi_{\delta}$ appear to have minimal impact on stPFI.

\begin{figure*}[h]
\centering
\includegraphics[width=\textwidth]{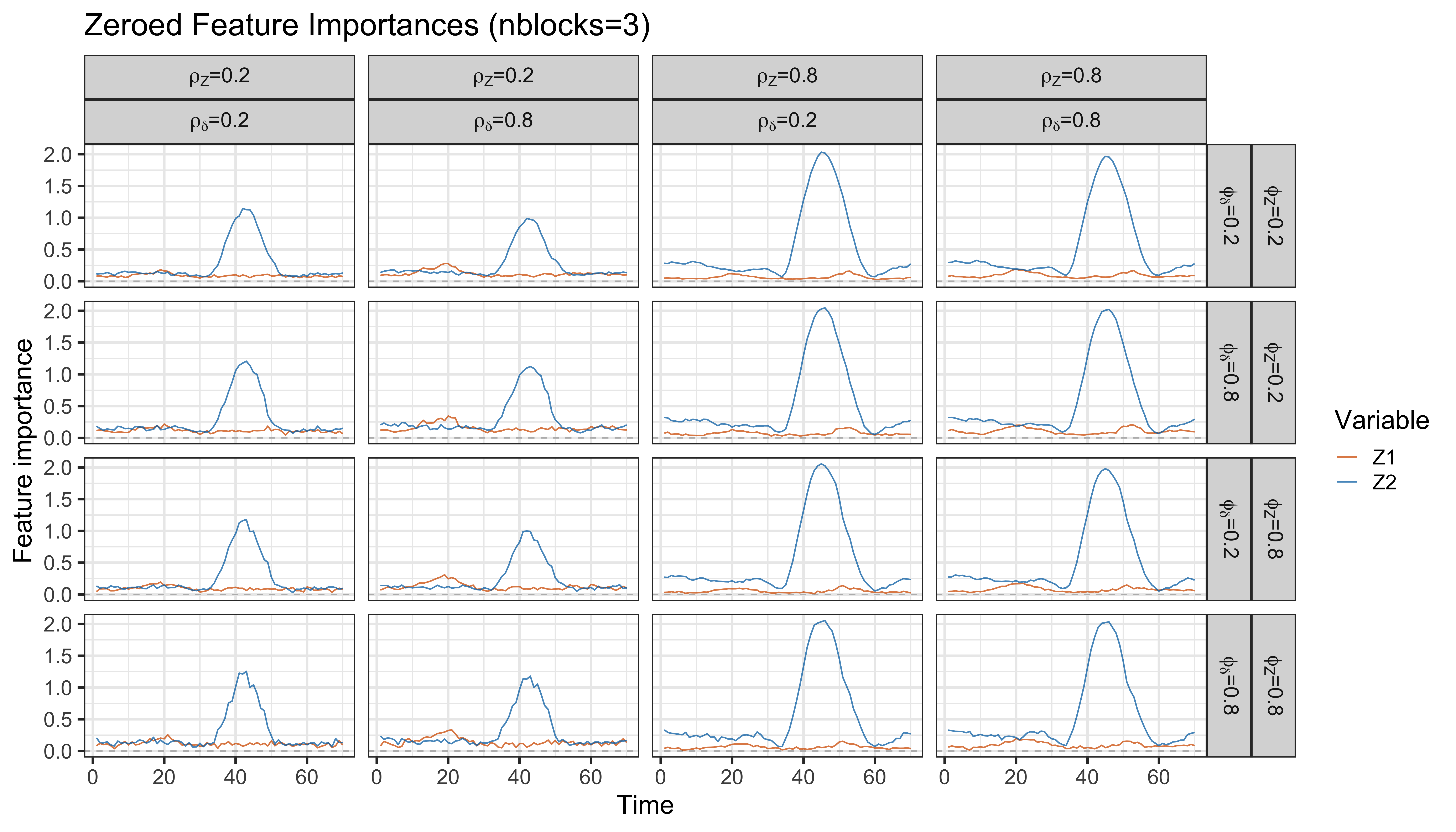}
\caption{Detailed simulation study results for stPFI and block size of 3. These plots explore changes to $\rho_z,\rho_{\delta},\phi_z,\phi_{\delta}$.}
\label{fig:zfi-supplement-rhophi}
\end{figure*}

Figure \ref{fig:zfi-supplement-rhophi} shows stZFI for variables ${\bf Z}_1$ and ${\bf Z}_2$, while varying $\rho_z,\rho_{\delta},\phi_z,\phi_{\delta}$ with a block size of 3. We restricted block size to 3 for brevity because there did not appear to be any interaction between block size and these parameters. The main difference is when $\rho_z$ is smaller, stZFI is relatively smaller for ${\bf Z}_2$ compared to when $\rho_z$ is large, meaning the performance of stZFI is better with larger $\rho_z$. Larger autocorrelation of the random effect, $\rho_{\delta}$, appears to make stZFI perform slightly worse (relatively smaller stZFI for ${\bf Z}_2$), but this effect appears small. The spatial range parameters, $\phi_z,\phi_{\delta}$ appear to have minimal impact on stZFI.

\subsection*{Mount Pinatubo Application Additional Results}

\begin{figure*}[h]
\centering
\includegraphics[width=\textwidth]{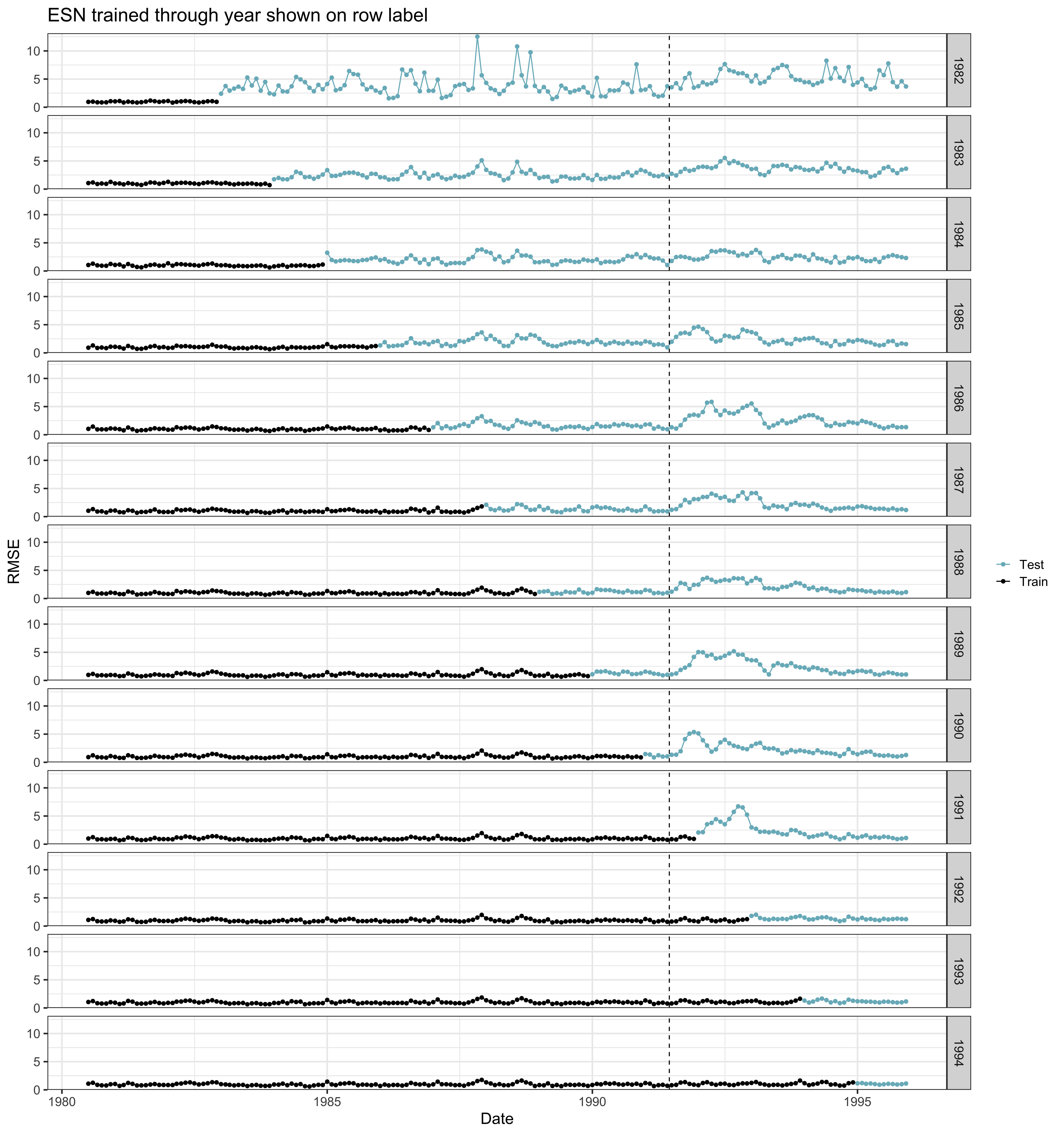}
\caption{RMSE for ESN on MERRA-2 data for training and testing splits. Each row represents an additional year of training data. Mount Pinatubo eruption is denoted by the vertical dashed line.}
\label{fig:merra2rmse}
\end{figure*}

Figure \ref{fig:merra2rmse} shows training and testing RMSEs for the ESN predicting stratospheric temperatures on MERRA-2. RMSEs are averaged over all spatial locations and plotted over time. The year in each row specifies through which year the model was trained on. Mount Pinatubo eruption is denoted by the vertical dashed line. When the model is trained with limited data (1980-1985), RMSE on the test set are large throughout. When enough training data is used, test RMSE before Mount Pinatubo eruption look decent, but unsurprisingly, the eruption causes poor RMSE performance since there is a change in the climate mechanism. However, when the ESN is trained through 1992 or beyond, test RMSE closely resemble training RMSE. The significant impact of Mount Pinatubo on the stratospheric temperatures in 1991 and beyond is clear.

\begin{figure*}[h]
\centering
\includegraphics[width=0.9\textwidth]{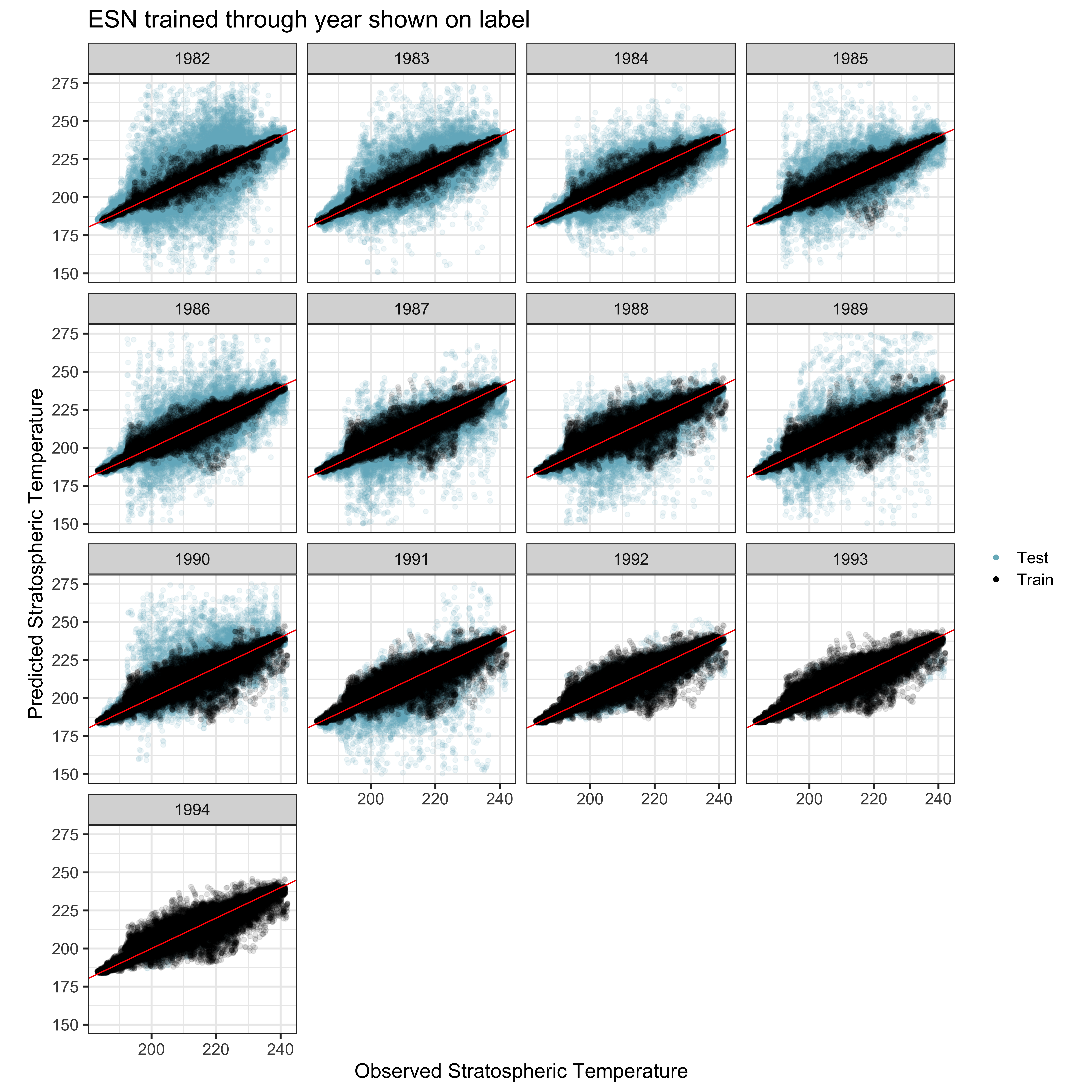}
\caption{Observed vs predicted stratospheric temperatures (on original scale) for ESN on MERRA-2 data for training and testing splits. The year in each row specifies through which year the model was trained on, with the remaining years through 1995 as test data. }
\label{fig:merra2pred}
\end{figure*}

Figure \ref{fig:merra2pred} shows observed vs predicted stratospheric temperatures (on original scale) for ESN on MERRA-2 data for training and testing splits. Each plot is labeled by the year through which the model was trained on, leaving the remaining years through 1995 as testing data. It is easy to visually see the effect of model underfitting in the first plots where there is significant variation in the test sets. Models trying to forecast past 1990 without having seen Mount Pinatubo struggle. Models trained through Mount Pinatubo and its aftermath (1992 and beyond) are able to predict into the future very well, even better than their training data. This is because the years following Pinatubo's eruption were less tumultuous. 

\begin{figure*}[h]
\centering
\includegraphics[width=\textwidth]{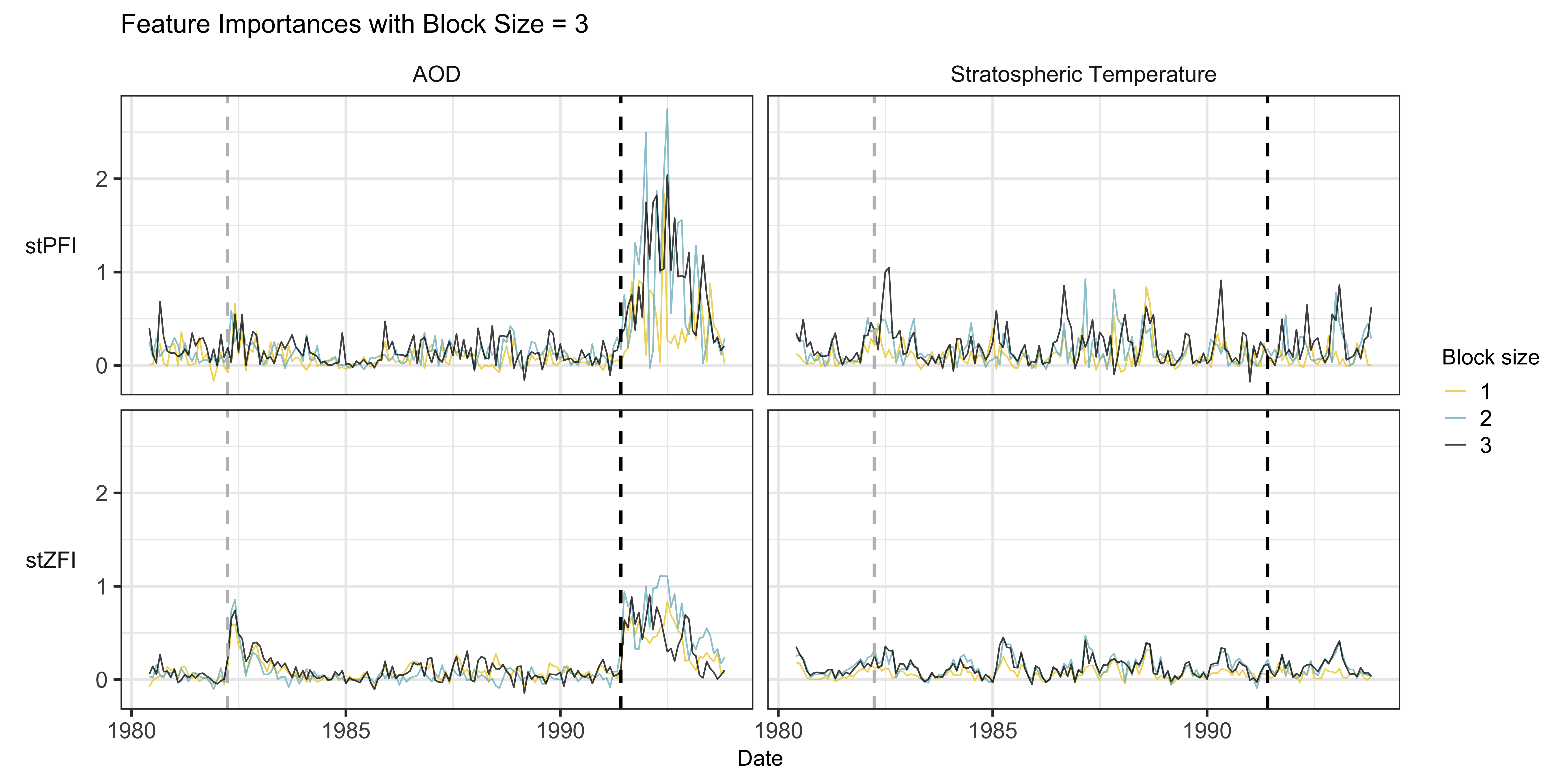}
\caption{Feature importances for different block sizes on MERRA-2.}
\label{fig:merra2block}
\end{figure*}

Figure \ref{fig:merra2block} shows FI on MERRA-2 data for different block sizes. The block size of three tends to be more stable, especially during the volcanic eruption events. Both stZFI and stPFI clearly pick up the importance of AOD for forecasting stratospheric temperatures, from Mount Pinatubo's eruption. It is interesting that stZFI also picks up the importance of AOD from El Chic\`{o}n's eruption, while stPFI does not. However, the simulation study in the main paper provided evidence that stZFI may be a better approach for detecting importance.

\end{document}